\documentclass[lettersize,journal]{IEEEtran}
\usepackage[
    colorlinks=true,
    linkcolor=orange,
    citecolor=blue,
    urlcolor=magenta]{hyperref}
\usepackage{amsmath,amsfonts}
\usepackage{amssymb}
\usepackage[capitalise, nameinlink]{cleveref}
\usepackage{algorithmic}
\usepackage{array}
\usepackage[caption=false,font=normalsize,labelfont=sf,textfont=sf]{subfig}
\usepackage{textcomp}
\usepackage{stfloats}
\usepackage{url}
\usepackage{verbatim}
\usepackage{graphicx}
\usepackage[table]{xcolor}
\usepackage{soul}
\usepackage{xspace}
\usepackage{booktabs}
\usepackage{float}
\usepackage{multirow}

\definecolor{RevisionColor}{RGB}{0,0,0}
\newcommand{\edit}[1]{#1}

\def\BibTeX{{\rm B\kern-.05em{\sc i\kern-.025em b}\kern-.08em
T\kern-.1667em\lower.7ex\hbox{E}\kern-.125emX}}
\usepackage{balance}
\begin{document}

\title{Clutter-\edit{Robust} Vision--Language--Action Models through Object-Centric and Geometry Grounding}
\author{Khoa~Vo\thanks{Khoa Vo is the corresponding author (e-mail: \texttt{khoavoho@uark.edu}).},
Taisei~Hanyu,
Yuki~Ikebe,
Trong~Thang~Pham,
Nhat~Chung,
Minh~Nhat~Vu,
Duy~Nguyen~Ho~Minh,
Anh~Nguyen,
Anthony~Gunderman,
Chase~Rainwater,
and Ngan~Le%
\thanks{Khoa Vo, Taisei Hanyu, Yuki Ikebe, Trong Thang Pham, Anthony Gunderman, Chase Rainwater, and Ngan Le are with the University of Arkansas, Fayetteville, AR, USA.}%
\thanks{Nhat Chung is with the National University of Singapore, Singapore.}%
\thanks{Minh Nhat Vu is with TU Wien, Vienna, Austria.}%
\thanks{Duy Nguyen Ho Minh is with Max Planck Research School for Intelligent Systems and the University of Stuttgart, Stuttgart, Germany.}%
\thanks{Anh Nguyen is with the University of Liverpool, Liverpool, U.K.}%

\small{\url{https://uark-aicv.github.io/OBEYED_VLA}}

}

\definecolor{Gray}{gray}{0.9}
\newcommand{\model}{OBEYED-VLA\xspace}
\newcommand{\modelPi}{OBEYED Pi-0\xspace}
\newcommand{\modelFAST}{OBEYED Pi-0 FAST\xspace}
\newcommand{\fullmodel}{OBject-centric and gEometrY groundED VLA\xspace}
\newcommand{\cmark}{\textcolor{green!55!black}{\checkmark}}
\newcommand{\xmark}{\textcolor{red!70!black}{$\times$}}
\renewcommand{\figurename}{Figure}


\maketitle

\begin{figure*}[t!]
    \centering
    \includegraphics[width=0.85\linewidth]{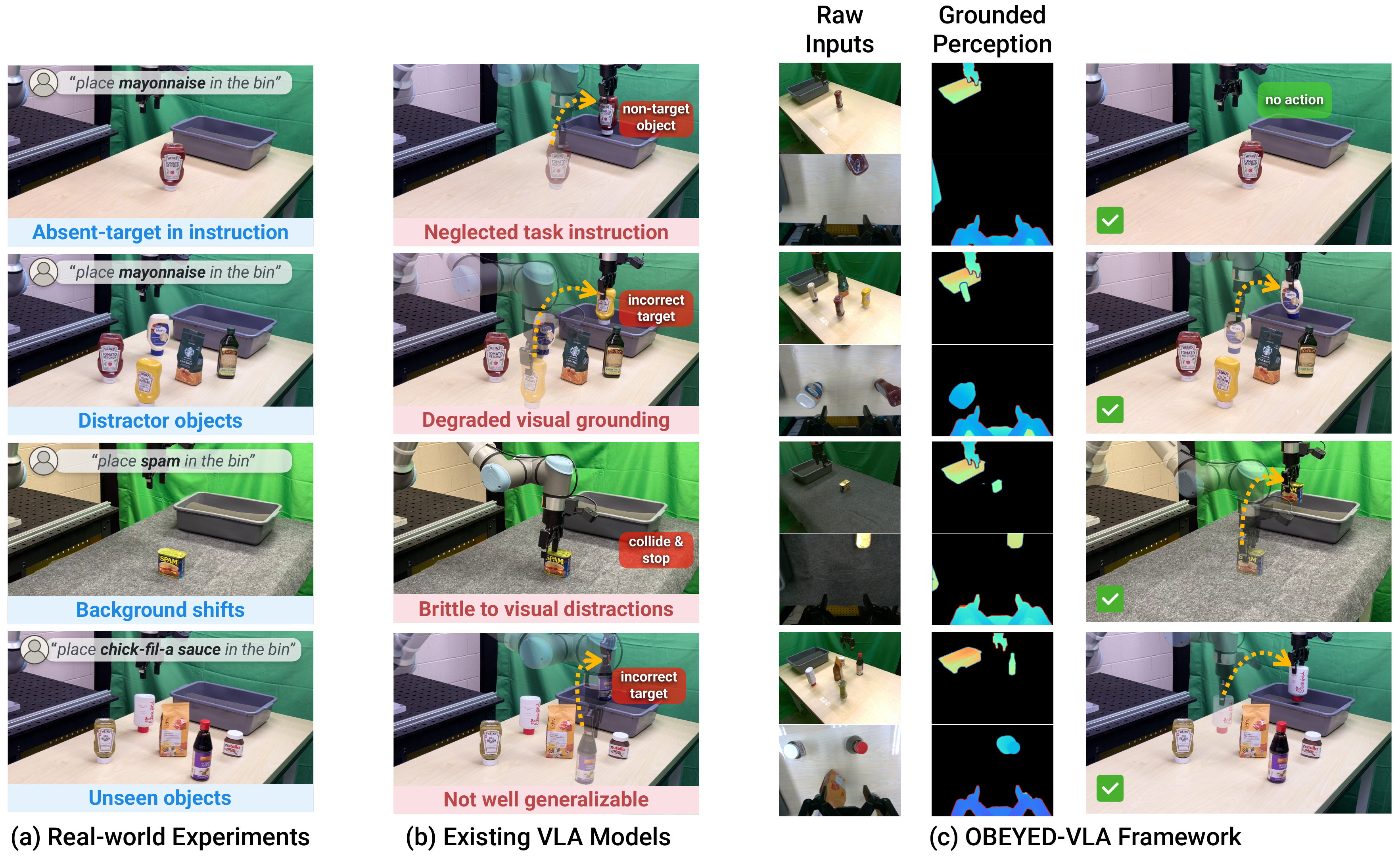}
    \caption{\textbf{Perception-grounded visuomotor manipulation in real-world cluttered scenes.} \textbf{(a) Real-world scenarios} that stress language-conditioned grounding, including mismatched task queries (absent targets),
    distractor objects, background appearance shifts, and unseen objects. \textbf{(b)~Typical~failure~modes of state-of-the-art VLAs}, which suffer degraded visual grounding, neglect task instructions, and are brittle to visual distractions, leading to spurious grasps, collisions, or picking incorrect targets. \textbf{(c) Proposed \fullmodel (\model) framework:} a VLM-driven perceptual module transforms raw RGB observations into task-conditioned, object- and geometry-focused views, enabling the downstream VLA to (i) remain reliable in cluttered scenes (e.g., with multiple distractor objects or shifted backgrounds), (ii) reject infeasible or inconsistent commands  and ignore distractors (e.g., absent-target instructions), and (iii) generalize to novel target objects unseen during training, without synthetic clutter data or auxiliary training losses.
    }
    \label{fig:train_test_setup}
\end{figure*}

\begin{abstract}
Recent Vision--Language--Action (VLA) models have made impressive progress toward general-purpose robotic manipulation by post-training large Vision--Language Models (VLMs) for action prediction. However, existing VLAs treat perception and control as a tightly entangled monolithic pipeline optimized purely for action. This end-to-end paradigm erodes language-conditioned visual grounding: in our real-world experiments, policies over-grasp when the requested object is absent, are easily distracted by clutter, and overfit to background appearance, leading to degraded performance in realistic tabletop scenes with distractors, background shifts, and unseen objects.

To address these issues, we propose \emph{\fullmodel} (\model), a framework that explicitly disentangles perceptual grounding from action reasoning. Instead of operating directly on raw RGB, \model augments VLAs with a perception module that grounds multi-view inputs into task-conditioned, object-centric, and geometry-aware observations. This module comprises a VLM-based object-centric grounding stage that selects task-relevant object regions across camera views, and a complementary geometric grounding stage that emphasizes the 3D structure of these objects over their appearance. The resulting grounded views are then fed to a pretrained VLA policy. We fine-tune this policy exclusively on single-object demonstrations captured without environmental clutter or non-target objects.

On a real-world UR10e tabletop setup, \model substantially improves robustness over strong VLA baselines across four challenging regimes and multiple levels of difficulty: distractor objects, absent-target rejection, background appearance changes, and cluttered manipulation of unseen objects. Ablation studies confirm that both object-centric and geometry-aware grounding are critical to these gains. Together, the results indicate that making perception an explicit, object-centric component is an effective way to strengthen and generalize VLA-based robotic manipulation.
\end{abstract}


\section{Introduction}

Recently, Vision-Language-Action (VLA) models such as Octo~\cite{octo_rss2024}, RoboFlamingo~\cite{roboflamingo_iclr2024}, OpenVLA~\cite{openvla_corl2024}, $\pi_0$~\cite{pi0_2024}, $\pi_0$-FAST~\cite{fast_2025}, and Gr00T~\cite{gr00t_2025} have made remarkable progress toward developing generalist visuomotor policies. These models unify vision, language, and robot control within a single framework that operates in two tightly coupled stages: (i) The \textit{perception stage} derives a semantic understanding of both the visual scene and the given instruction, while (ii) the \textit{action reasoning stage} builds upon this understanding to generate executable control sequences. Notably, through large-scale pretraining on diverse robot demonstration datasets (e.g., BridgeData V2~\cite{bridgedatav2_corl2023}, OXE~\cite{oxe_corl2024}, DROID~\cite{droid_2024}, and $\pi$ dataset~\cite{pi0_2024}) that encompass a wide range of manipulation tasks, VLAs exhibit promising transferable action reasoning capabilities that enable them to adapt effectively to novel downstream tasks and embodiments.

Despite their promising transferability in action reasoning, VLAs are bottlenecked by perception stage, a \emph{reliable language-conditioned visual grounding} often collapses in real-world \emph{cluttered scenarios}. In our real-world experiments as shown in \cref{fig:train_test_setup}~(a \& b), we observe these
failure modes of existing VLAs: the policy often misaligns referring expressions with the correct target, latches onto task-irrelevant distractors, or executes an action even when the instruction is inconsistent with the scene, indicating that linguistic cues are not consistently tied to the right visual evidence.

We attribute this brittleness to the prevailing VLA training paradigm in which perception and control are optimized end-to-end for action prediction. However, minimizing an action-centric objective does not, by itself, preserve stable object-level language–vision alignment. In particular, when fine-tuning data exhibits limited clutter variability and lacks hard negative cases (e.g., absent-target instructions), the model can achieve high training likelihood by learning shortcuts--such as object-presence priors that favor executing a grasp whenever a salient object is visible, or reliance on background and context-specific cues. Consequently, the vision–language representations that VLAs inherit from pretrained VLM backbones may drift toward action-effective but grounding-weak features, which manifests as over-grasping, distractor sensitivity, and poor robustness under clutter and distribution shift.





While scaling up downstream datasets with synthetic cluttered scenes or introducing auxiliary perceptual objectives (e.g., ECoT~\cite{ecot_2024}, FAST-ECoT~\cite{fast_ecot_2025}, and CoT-VLA~\cite{cot_vla_cvpr2025}) can partially alleviate these issues, such approaches demand prohibitively extensive effort in data collection, and annotation. In addition, a larger dataset would significantly prolong training, thus increases computational cost. These challenges motivate a central question:
\emph{Without relying on synthetic cluttered data or additional perceptual objectives, can we strengthen the perception ability of a VLA model so that it remains reliable in clutter, robust against distractors, and able to generalize to unseen objects?}


\edit{Vision-Language Models (VLMs), such as GPT-4V~\cite{gpt4_arxiv2024}, BLIP-2~\cite{blip2_icml2023}, Qwen2.5-VL~\cite{qwen25-vl}, and Qwen3-VL~\cite{qwen3-vl}, are trained on web-scale image--text datasets and exhibit strong language-conditioned visual grounding in zero-shot settings when equipped with visual prompting strategies such as set-of-mark~\cite{setofmark_2023}. This capability has motivated a growing line of hierarchical robotic manipulation systems in which a high-level VLM or Multimodal Large Language Model (MLLM) generates task-relevant intermediate representations for downstream control. In RoboGround~\cite{roboground_arxiv2025}, CrayonRobo~\cite{crayonrobo_arxiv2025}, ManipLLM~\cite{manipllm_arxiv2024}, HAMSTER~\cite{hamster_iclr2025}, MOKA~\cite{moka_rss2024}, and ReKep~\cite{rekep_corl2025}, the high-level module predicts task-specific masks, prompts, contact poses, 2D end-effector paths, keypoints, or constraints that are then consumed by downstream policies or low-level robot controllers. In HiRobot~\cite{hirobot_icml2025}, a VLM is trained to produce language subtask commands for a downstream VLA in long-horizon tasks.
These approaches demonstrate the promise of decoupling perception from control. However, several of them require extensive, task-specific high-level training data or supervision, reintroducing the data-collection and annotation burden discussed above. In addition, they are designed for generalizable, open-world, or long-horizon manipulation while assuming largely clutter-free scenes. Recently, BYOVLA~\cite{byovla_icra2025} directly addresses scene distractions via test-time observation editing: it localizes distractors and removes them before passing the edited observations to a pretrained, frozen VLA for action prediction. However, this intervention pipeline is computationally heavy, requiring repeated VLA forward passes to estimate distractor-sensitive regions and a diffusion-based inpainting step during inference.
}

\edit{To address this gap, we present \textbf{\fullmodel (\model)}, a hierarchical framework that improves language-conditioned visual grounding for VLAs in cluttered scenes. \model requires no additional high-level training. Instead, it uses off-the-shelf VLM grounding and explicit cross-view reasoning to transform cluttered multi-view observations into task-conditioned inputs for the VLA that preserve the target object and its relevant geometry while suppressing distractors. The VLA then reasons over clutter-free, geometry-aware observations instead of raw cluttered images (\cref{fig:train_test_setup}~(c)).}


\edit{Specifically, \model comprises two coupled grounding stages. First, an object-centric grounding stage uses an off-the-shelf VLM to identify instruction-relevant regions and match them across camera views, suppressing distractors while keeping the target object visible to the VLA. Second, a geometric grounding stage converts the selected regions into depth-based observations that emphasize 3D structure over appearance. The downstream VLA is then fine-tuned to perform action reasoning on these grounded observations using only clean single-object demonstrations, without synthetic clutter augmentation or auxiliary perceptual objectives. Unlike BYOVLA, our method simply masks out distractor regions directly in the VLA inputs rather than removing them through expensive multiple forward passes and diffusion-based inpainting, making inference substantially lighter.}

Overall, our key contributions are summarized as follows:
\begin{itemize}
    \item We propose \model, a \edit{hierarchical} framework that equips VLAs with object geometry grounding, providing semantically relevant and spatially grounded observations to VLAs for visuomotor reasoning. \item Through extensive real-world experiments, \model shows superior robustness to cluttered scenes with various distraction settings and environmental clutter challenges compared to strong VLA baselines, despite fine-tuning only on clean single-object demonstrations.
    \item We show that \model can effectively generalize to unseen target objects with novel scene compositions, maintaining reliable visuomotor performance.
\end{itemize}

\section{Related Work}


\begin{table*}[t]
\centering
\scriptsize
\setlength{\tabcolsep}{2pt}
\renewcommand{\arraystretch}{1.2}
\caption{\textcolor{RevisionColor}{Comparison of hierarchical perception-control methods for robotic manipulation.
}}
\label{tab:related_work_comparison}
\resizebox{\textwidth}{!}{%
{\color{RevisionColor}\arrayrulecolor{RevisionColor}%
\begin{tabular}{@{}l|ccc|cc|cc|c@{}}
\toprule
\multirow{2}{*}{\textbf{Method}} &
  \textbf{High-level} & \textbf{High-level} & \textbf{High-level} &
  \textbf{Low-level} & \textbf{Low-level} &
  \textbf{Cross-view} & \textbf{Clutter} & \textbf{Target Problem} \\
  &
  \textbf{type} & \textbf{output} & \textbf{training data} &
  \textbf{type} & \textbf{training data} &
  \textbf{grounding} & \textbf{handling} & \textbf{(robotic manip.)} \\
\midrule
RoboGround~\cite{roboground_arxiv2025} & VLM & Target/place masks & Sim QA+mask pairs & Transformer policy & Robot demos & \xmark & \xmark & Generalizable \\
CrayonRobo~\cite{crayonrobo_arxiv2025} & VLM & Visual prompts & None & MLLM & Sim prompt-action & \xmark & \xmark & Generalizable \\
ManipLLM~\cite{manipllm_arxiv2024} & MLLM & Contact pose & Sim affordance/pose & Pose controller & None & \xmark & \xmark & Generalizable \\
HiRobot~\cite{hirobot_icml2025} & VLM & Subtask commands & Real/synthetic prompts & VLA & Robot demos & \xmark & \xmark & Long-horizon\\
HAMSTER~\cite{hamster_iclr2025} & VLM & 2D end-effector path &  Real/sim 2D path+VQA & 3D policy & Robot demos & \xmark & \xmark & Open-world \\
MOKA~\cite{moka_rss2024} & VLM & Keypoint affordances & None & Pose controller & None & \xmark & \xmark & Open-world \\
ReKep~\cite{rekep_corl2025} & VLM & Keypoint constraints & None & Optimization solver & None & \xmark & \xmark & Open-world \\
BYOVLA~\cite{byovla_icra2025} & VLM & Clutter-free obs. & None & VLA & None & \xmark & Inpaint & Clutter-robust \\
\rowcolor{Gray}
\textbf{\model} & VLM & Clutter-free geom. obs. & None & VLA & Robot demos & \cmark & Mask-out & Clutter-robust  \\
\bottomrule
\end{tabular}
}%
}
\end{table*}
\arrayrulecolor{black}

\subsection{Vision-Language-Action (VLA) models}
\label{sec:related_vla}
Building upon the success of VLMs in cross-modal understanding tasks~\cite{paligemma_2024, prismatic_icml2024, qwen25-vl}, recent years have seen the emergence of VLAs that demonstrate strong generalization in robotic control \cite{openvla_corl2024, pi0_2024, gr00t_2025, fast_2025, cot_vla_cvpr2025}. The central idea of VLAs is to transfer rich semantic and perceptual knowledge learned by pretrained VLMs into visuomotor policy learning. VLAs post-train VLM backbones on action prediction tasks, leveraging large-scale robot datasets~\cite{bridgedatav2_corl2023, droid_2024, oxe_corl2024, pi0_2024}, which span a broad range of manipulation skills and robot embodiments.

Two primary architectures are commonly employed for action prediction. \emph{Autoregressive VLAs}~\cite{openvla_corl2024, fast_2025} discretize robot actions into tokens and formulate robot control as a next-token prediction problem, enabling direct transfer of semantic reasoning in VLMs to embodied control. In contrast, \emph{flow-based VLAs}~\cite{pi0_2024, pi05_2025, gr00t_2025} generate continuous actions by transforming noise into action trajectories via learned continuous-time dynamics (e.g., flow matching~\cite{flowmatching_iclr2023}), offering smoother and higher-frequency control.

Although these models show promising transferability in action reasoning, being optimized exclusively
with robot control objectives leads to a degradation of visual--language perception, reducing robustness to distractors, cluttered scenes, and instruction following. As illustrated in \cref{fig:train_test_setup}~(b), VLAs are easily distracted by irrelevant objects, often fail to tolerate background changes, and struggle to associate referring instruction with the correct target in cluttered or novel-object scenes. Several works attempt to alleviate these issues by introducing auxiliary perception-focused objectives, such as visual reconstruction losses, spatial grounding losses, or contrastive vision-language alignment terms~\cite{ecot_2024, fast_ecot_2025, cot_vla_cvpr2025}. Other approaches instead co-train on both vision--language reasoning data and robot control demonstrations~\cite{pi05_2025, gr00t_2025}, which strengthen visual--language grounding but require substantial additional data and computation.


Crucially, these approaches still implement perception and action prediction within a monolithic architecture that is optimized end-to-end. On their original training domains, optimizing this unified model with a combination of action-prediction and auxiliary perception-oriented objectives, supported by rich annotations, can maintain strong vision--language alignment. However, when the same architecture is adapted to new tasks, embodiments, or environments, downstream datasets typically lack the supervision needed to sustain these auxiliary objectives. Fine-tuning collapses to solely action prediction loss, which again erodes vision--language alignment. Retaining the auxiliary perception objectives during adaptation would require collecting additional perceptual labels for every new deployment setting, which is rarely practical. As a result, such monolithic pretraining schemes are poorly suited to preserving reliable language-conditioned visual grounding in VLA policies under downstream adaptation.

\emph{Our proposed framework instead explicitly decouples perception from control by augmenting existing VLAs with a dedicated perception grounding module. This module operates on raw observations to produce semantically and spatially focused inputs—suppressing irrelevant regions, isolating task-relevant objects, and emphasizing object-centric geometry—before they are passed to the VLA policy for action reasoning. By separating perception grounding from action prediction, we improve robustness to various scene clutter types and generalization to unseen objects without requiring additional cluttered demonstrations or auxiliary perceptual objectives during VLA training, and we can reuse the same perception module across different VLAs, environments, and robot embodiments.}


\subsection{Vision-Language Models (VLMs) as high-level perception experts}
\edit{Pretrained on internet-scale image-text data, VLMs demonstrate strong semantic understanding and generalization~\cite{qwen25-vl, qwen3-vl, gpt4_arxiv2024}. These strengths have driven a growing effort in robotics to employ VLMs as high-level perception and reasoning modules for control policies.}

\edit{Some recent works only employ VLMs as external monitors around an existing policy. AHA~\cite{aha_2024} and FailSafe~\cite{failsafe_2025} fine-tune VLMs to monitor robot behavior, detect failures, and generate corrective interventions that override a base policy when necessary. In these systems, the high-level VLM does not communicate continuously with the low-level policy, but intervenes externally only when failures are detected. While these systems enhance reliability, they require extensive task-specific VLM fine-tuning and rely on external intervention rather than directly improving the visual grounding of the underlying VLA policy.}

\edit{\cref{tab:related_work_comparison} compares representative hierarchical perception-control methods for robotic manipulation, including RoboGround~\cite{roboground_arxiv2025}, CrayonRobo~\cite{crayonrobo_arxiv2025}, ManipLLM~\cite{manipllm_arxiv2024}, HAMSTER~\cite{hamster_iclr2025}, MOKA~\cite{moka_rss2024}, ReKep~\cite{rekep_corl2025}, and HiRobot~\cite{hirobot_icml2025}. It highlights their differences in high-level outputs, training data, clutter handling, and target problem. Although all of these methods decouple high-level reasoning from low-level execution, the interface between the two levels varies widely. RoboGround uses simulated question-answering (QA) and paired target-placement masks to train a transformer-based robot policy. CrayonRobo and ManipLLM instead rely on prompt- or pose-centric representations that are executed by MLLMs or pose controllers. HAMSTER uses 2D end-effector paths together with visual question answering (VQA) data to guide a compact 3D policy, whereas MOKA and ReKep use keypoint affordances or constraints to guide downstream motion planning and control. HiRobot trains a high-level VLM to issue language subtask commands to a downstream VLA for long-horizon execution. Several of these methods also require additional high-level training data or supervision. More importantly, they generally do not study cluttered tabletop scenes, but instead focusing on generalizable, open-world, or long-horizon manipulation under largely clutter-free conditions.}

\edit{Most recently, BYOVLA~\cite{byovla_icra2025} improves VLA robustness by performing observation interventions during inference. It first queries a VLM to identify distractor objects and localize their regions using a segmentation model. Next, it employs GradCAM~\cite{gradcam_ijcv2020} to determine sensitive regions, and finally removes those regions by a diffusion-based inpainting model. Although this process enhances performance in cluttered scenes, it introduces significant computational overhead. Each distractor region requires an additional VLA forward pass to test that region is sensitivity-relevant to the VLA, and the final edited observation is then produced through an expensive inpainting step, making real-time operation impractical.}

\edit{\emph{Our proposed framework instead uses an off-the-shelf VLM to ground multi-view observations into object-centric RGB and geometry-aware inputs for a downstream VLA, without additional high-level training. Compared with BYOVLA, the closest prior work in this setting, our method resolves clutter through explicit cross-view grounding and direct distractor suppression rather than expensive single-view test-time editing and inpainting.}}

\begin{figure*}[t]
    \centering
    \includegraphics[width=\linewidth]{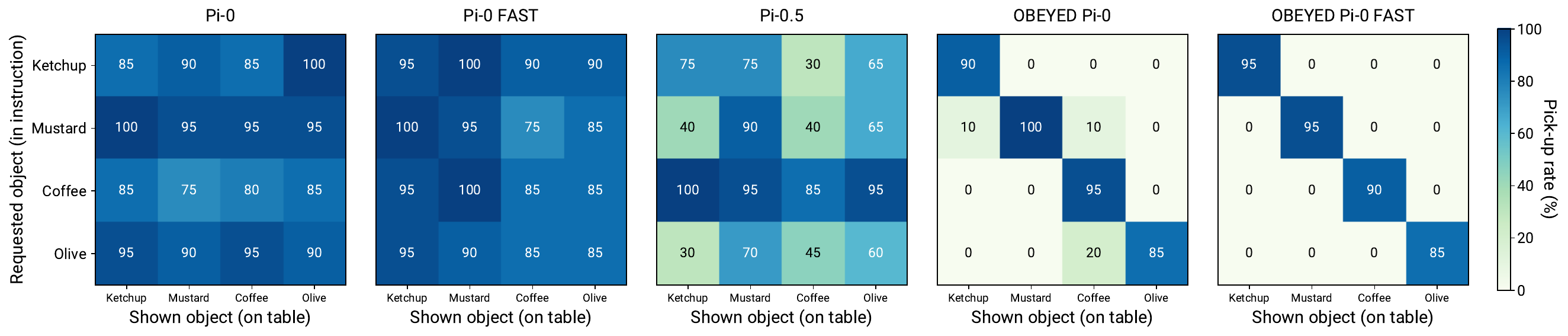}
    \caption{\textbf{Absent-target sanity check of vision-language grounding.} We report pick-up rate (\%) for each (requested, shown) object pair, computed over 20 rollouts for all combinations of requested (rows) and shown (columns) objects. Object labels are \emph{Ketchup}, \emph{Mustard}, \emph{Coffee} (coffee bag), and \emph{Olive} (olive oil bottle), so off-diagonal intensities directly reveal how often the policy grasps when the requested object is absent.}
    \label{fig:neg_prompt_heatmaps}
\end{figure*}

\section{Preliminary \& Problem Statement}
\label{sec:prelim}
\textbf{Preliminary.}
We formulate robotic manipulation through a visuomotor policy \( \pi_\theta \),
which predicts a short-horizon action trajectory of size $H$:
\begin{equation}
\tau_t = (a_t,\dots,a_{t+H}) \sim \pi_{\theta}(o_t, q_t, l)
\end{equation}
given a natural language instruction $l$, visual observation $o_t$, and a robot's proprioceptive state $q_t$ at a timestep $t$. For clarity of exposition, we omit timestep subscript in the remainder of this paper.

In this work, our robot setup (\cref{fig:experimental_setup}) provides each observation as two RGB inputs from distinct viewpoints: an over-the-shoulder camera mounted on the robot base, providing $I^{base}$, and a wrist-mounted camera that captures $I^{wrist}$. We denote the combined visual observation as
\(o=(I^{base}, I^{wrist})\), though additional camera views can be incorporated seamlessly in future extensions.

The policy \(\pi_\theta\) is trained on a dataset of robot demonstrations, each of which is decomposed into a sequence of frame-wise samples, where a sample \(i^{th}\) contains a visual observation \(o_i\),
the corresponding proprioceptive state \(q_i\), a short-horizon action segment \(\tau_i\), and the associated language instruction \(l_i\). These samples form the dataset:
\begin{equation}
\mathcal{D}=\{(o_i, q_i, \tau_i, l_i)\}_{i=1}^{N}
\end{equation}
and the policy is optimized via maximum-likelihood estimation to match the demonstrated action sequences:
\begin{equation}
\label{eq:optimization}
\max_\theta \mathbb{E}_{(o, q, \tau, l)\sim \mathcal{D}}\left[\log \pi_\theta (\tau|o,q,l)\right]
\end{equation}

\textbf{Problems of baselines.}
In practice, a pretrained VLA model can be adopted for visuomotor policy \(\pi_\theta\) and fine-tuned on \(\mathcal{D}\).
Owing to their large-scale pretraining on diverse robot datasets, such models adapt their action distributions to new embodiments and workspaces with relatively modest amounts of downstream data.
However, because perception and action reasoning are tightly coupled and optimized end-to-end solely for action prediction, the vision–language alignment inherited from the underlying VLM backbones is gradually distorted by the control objective, weakening language-conditioned visual grounding.

We investigate this misalignment explicitly through a simple absent-target sanity check summarized in \cref{fig:neg_prompt_heatmaps}.
In this experiment, we place a single object (e.g., ketchup) on the table and issue either a matching instruction (e.g., \textit{``place ketchup in the bin''}) or a mismatched instruction that refers to a different object (e.g., \textit{``place mustard in the bin''}).
The correct behavior is straightforward: the policy should pick up the object only when the instruction matches the physical object shown on the table, and otherwise refrain from grasping.
For each pair of \emph{requested} object and \emph{shown} object, we measure the empirical pick-up rate, yielding a heatmap in which a well-grounded policy would have high values only on the diagonal and near-zero values elsewhere.

The heatmaps in \cref{fig:neg_prompt_heatmaps} show that Pi-0, Pi-0 FAST, and Pi-0.5 systematically violate this basic behavior. Across almost all off-diagonal entries, where the requested object is absent, their pick-up rates remain high, often above 75\%, indicating that these policies give little weight to the linguistic command and instead default to executing a grasp whenever a plausible object is present in the scene.

This experiment, conducted in the simplest single-object setting without clutter, highlights the fundamental limitation of current VLAs: monolithic end-to-end action fine-tuning encourages almost unconditional grasping behavior and progressively erodes the underlying vision–language alignment, leading to poor language-conditioned visual grounding. Full training configurations and a comprehensive quantitative comparison of these baselines are provided in \cref{sec:experiments}.


\textbf{Our objective.}
We aim to strengthen the perception capability of VLA policies. We study this problem in a tabletop pick-and-place setting where the training demonstrations in $\mathcal{D}$ contain only clean and single-object scenes. To systematically probe robustness at deployment, we consider four evaluation scenarios (\cref{fig:train_test_setup}~(a)): (i) cluttered scenes with distractor objects, queried either by object identity or by spatial reference; (ii) absent-target instructions that require the policy to abstain when the queried object is missing; (iii) distribution shifts in background appearance; and (iv) manipulation of novel, previously unseen objects. Together, these scenarios test whether a policy preserves reliable language-conditioned visual grounding beyond the narrow training distribution.

\begin{figure*}[t]
    \centering
    \includegraphics[width=\linewidth]{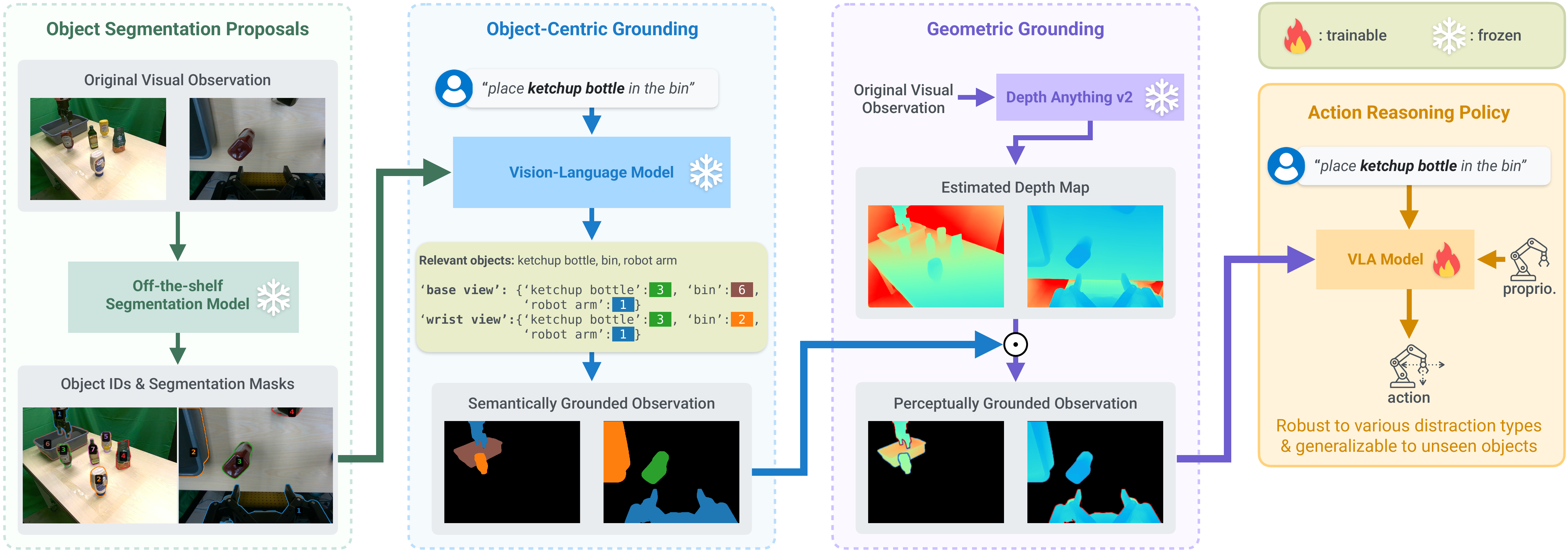}
    \caption{\textbf{An overview of \model architecture.}
Raw RGB images from base and wrist cameras are first passed through a segmentation network to obtain object-level masks. VLM-based \textbf{object-centric grounding module} then selects a subset of masks corresponding to task-relevant objects, while \textbf{geometric grounding module} applies depth estimation to these masks to produce clutter-suppressed, geometry-aware observations focused on those regions. The resulting perceptually grounded observations, together with the language instruction and robot proprioception, are then fed into a pretrained VLA model that outputs action trajectories; only the VLA needs to be fine-tuned for downstream tasks, while the perception modules remain frozen to enable plug-and-play integration with different VLAs.
}
    \label{fig:arch}
\end{figure*}

\section{\model}
In this section, we present the \fullmodel (\model) for the language-conditioned robotic manipulation. At a high level, \model introduces a Perception Grounding module that grounds raw visual observations and transforms them into clutter-suppressed, geometry-aware visual inputs for the Vision-Language-Action model (VLA). Our aim is to improve performance of the VLA model on fine-grained task instructions in dense, distractor-heavy scenes and in scenarios involving novel target objects. In this section, we first describe the grounding module and its outputs in detail (\cref{sec:method_pg}), then explain how these outputs are integrated with arbitrary VLA models (\cref{sec:vla}).

\subsection{Perception Grounding Module}
\label{sec:method_pg}
Given the visual observation, our approach first generates mask proposals for all present objects in the workspace. These masks serve as visual prompts~\cite{setofmark_2023} for a VLM, which grounds the original visual observation to select regions that are most relevant to the task instruction.
For the selected regions, we first suppress all background along with irrelevant objects in the RGB image, yielding a view in which only instruction-relevant objects remain visible to the downstream VLA and thus concentrating its action reasoning on task-relevant content. We then convert the remaining pixels into a depth representation, preserving the 3D shape and spatial layout of the selected objects while discarding appearance cues such as color and texture, which encourages the policy to rely on geometry rather than superficial visual correlations. The overall perception grounding pipeline is illustrated in \cref{fig:arch}, and we detail each step below.

\textbf{Object Segmentation Proposals.}
We employ an off-the-shelf segmentation model to process the RGB observation from both camera views, \(I^{base}\) and \(I^{wrist}\), and produce object mask proposals \(\mathcal{M}^{base}=\{m^{base}_k\}_{k=1}^{K_{base}}\) and \(\mathcal{M}^{wrist}=\{m_k^{wrist}\}_{k=1}^{K_{wrist}}\) covering visible objects in the workspace, where \(K_{base}\) and \(K_{wrist}\) denote the numbers of detected objects in base and wrist views, respectively.
Each mask \(m_k^{\{base,wrist\}}\) defines a candidate object region in the corresponding view that will later be converted into a mark-based visual prompt for the object-centric grounding step.

One may apply open-vocabulary segmentation models in SAM family~\cite{sam_cvpr2023, sam2_iclr2025, semantic_sam_eccv2024, seem_neurips2023} to this task; however, these models often over-partition objects into multiple disjoint fragments owing to the nature of its pretrained datasets, i.e., SA-1B~\cite{sam_cvpr2023}. Consequently, the VLM is forced to infer which fragments belong together, introducing unnecessary reasoning overhead and frequently leading to incorrect grounding. Closed-vocabulary but large-coverage models like Co-DETR~\cite{co-detr_cvpr2023} trained on Objects365~\cite{objects365_cvpr2019}+LVIS~\cite{lvis_cvpr2019}, produce more coherent whole-object masks, yet they are not trained for robot arm and gripper segmentation, causing unreliable masks produced for these categories. Both SAM-based and Co-DETR models also have substantial computational cost, making them impractical for real-time deployment within a closed-loop manipulation system.

To unify the advantages of both worlds in a single efficient model, we fine-tune YOLO11-Seg~\cite{yolo11_2024} on a hybrid dataset that combines our robot demonstrations with a curated subset of LVIS~\cite{lvis_cvpr2019}. We first automatically annotate 100 teleoperated demonstrations using a unified pipeline that integrates both Co-DETR and SAM-based methods: workspace objects are annotated using whole-object masks from Co-DETR~\cite{co-detr_cvpr2023}, while the robot arm and gripper are localized with Grounding DINO~\cite{gdino_eccv2024} on the initial frame, segmented with SAM~\cite{sam_cvpr2023}, and then temporally propagated with Cutie~\cite{cutie_cvpr2024}. To improve coverage beyond our eight grocery objects, we additionally construct an LVIS subset by selecting categories corresponding to indoor tabletop items (e.g., bottles, cans, boxes, cups, and utensils) and retaining images that only contain such instances. YOLO11-Seg is then fine-tuned on a 50:50 mixture of the annotated demonstrations and this LVIS subset. This mixed training regime yields a segmentation module that reliably identifies diverse tabletop objects and the robot arm while supporting the real-time operation required by our manipulation system. \edit{Please refer to Appendix~\ref{app:yolo_training} for additional training details.}

\textbf{Object-Centric Grounding.}
Humans naturally perceive scenes through an object-centric lens. For instance, when asked to \textit{``place the ketchup bottle in the bin"}, we localize the ketchup bottle, the bin, and our hand in order to carry out the task. The presence of many other objects on the table has minimal influence on this perception. Irrelevant objects and background simply recede from attention, and focus narrows to the entities that matter for executing the instructed action. This object-centric perception allows humans to act reliably even in visually dense and cluttered environments.
Inspired by this intuition, our approach employs VLMs, specifically Qwen3-VL~\cite{qwen3-vl}, leveraging its emergent visual perception and reasoning, to ground the visual observations and isolate the regions most relevant to the given instruction.

\begin{figure}[t]
    \centering
    \includegraphics[width=\linewidth]{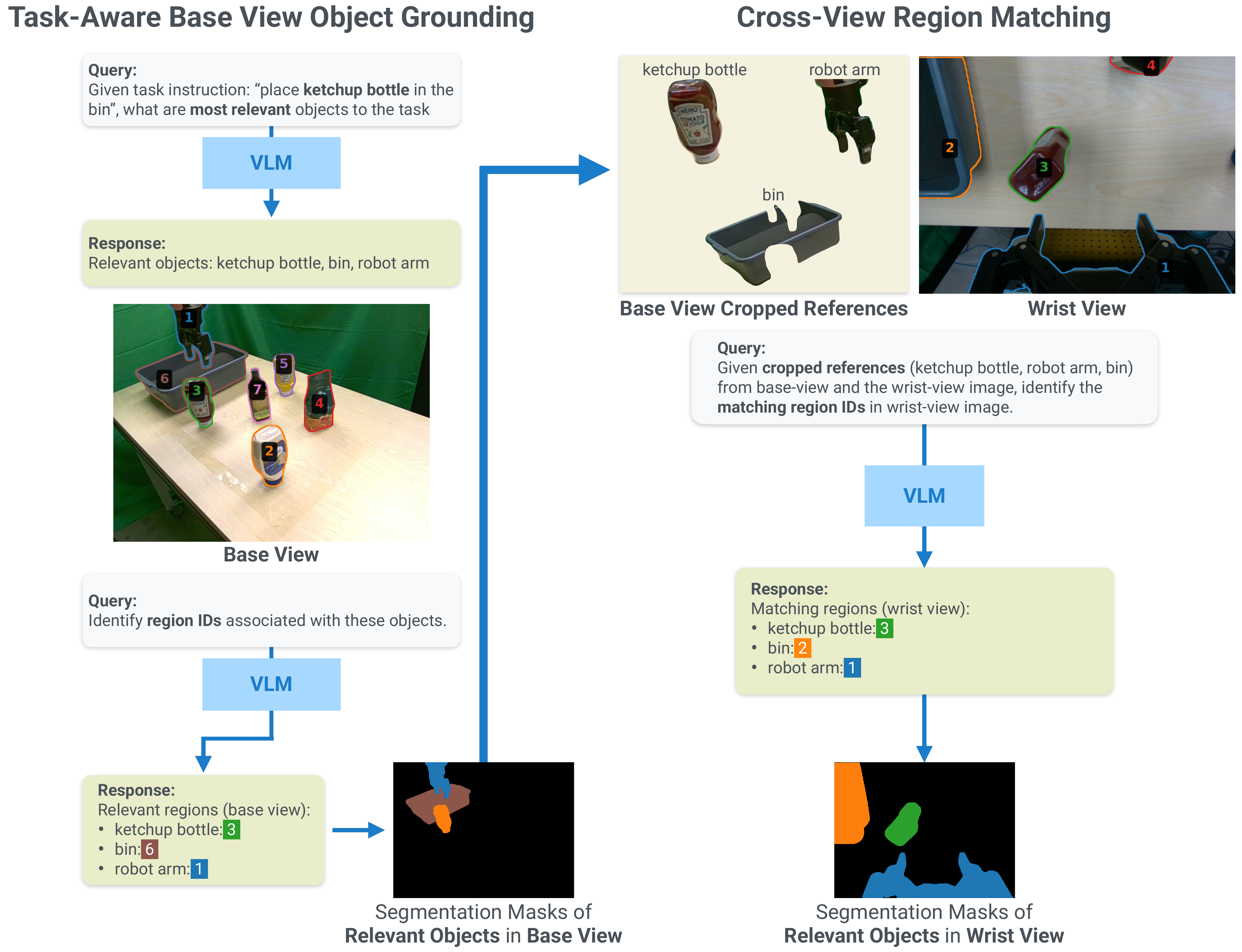}
    \caption{\textbf{Object-Centric Grounding Module.}
    The module operates in two stages. First, the VLM parses the task instruction to extract task-relevant objects and, using set-of-mark prompting on the base-view segmentation masks to select the regions corresponding to those objects. We crop the selected base-view regions to produce object-centric reference views and provide these, together with set-of-mark augmented wrist-view image, in a single prompt to the VLM, which predicts the matching wrist regions. The resulting task-relevant masks in both base and wrist views define semantically grounded regions that eliminate distractions and background, isolating only the visual content most relevant to the task instruction.
    }
    \label{fig:semantic_grounding}
\end{figure}

Our approach is designed as a two-stage object-centric grounding process: \textit{task-aware base-view object grounding} followed by \textit{cross-view region matching}, as shown in \cref{fig:semantic_grounding}.

\underline{\emph{Task-aware base-view object grounding.}}
We first perform a language-only parsing step on the task instruction \(l\) by prompting the VLM to list objects involved to fulfill the instruction (e.g., the queried object and the receptacle), yielding a set of object names \(\mathcal{E}(l)=\{e_j\}\) relevant to the task instruction.

Afterwards, given the base-view image and segmentation proposals \(\mathcal{M}^{base}=\{m_k^{base}\}\) that cover all candidate objects in the scene, we employ the set-of-mark visual prompting mechanism~\cite{setofmark_2023} by overlaying a \emph{numeric mark} (a positive number) inside each mask region \(m_k^{base}\) on top of the original RGB image. This produces a mark-augmented base-view image in which every segmented region is tagged with a distinct, spatially localized symbol. Overlaying markers directly onto the RGB image makes these identifiers visually aligned with the underlying region, providing explicit spatial references that help the VLM reason about individual regions.

We then query the VLM with object names \(\mathcal{E}(l)\) and mark-augmented base-view image. The model is prompted to identify which markers correspond to the task-relevant objects, producing a subset of masks
\begin{equation}
    \mathcal{S}^{base} \subseteq \mathcal{M}^{base}
\end{equation}
that it deems relevant to the instruction. Since the scenes in our experiments are largely static aside from the robot arm and the actively manipulated object, we invoke the VLM for this stage only once at the beginning of each rollout and then track the selected masks across the remaining frames.

For each \(m_k^{base} \in \mathcal{S}^{base}\), we further extract a tight RGB crop around the mask from the original base-view image and apply the corresponding binary mask within this cropped window to suppress background. This yields object-centric reference views in which only the selected object remains visible while surrounding clutter is removed, providing canonical visual anchors for the subsequent cross-view matching stage.


\underline{\emph{Cross-view region matching.}}
Observations from wrist-view often depict objects from top-down or oblique angles, where object appearances deviate substantially from the typical front-view, object-upright that VLMs mostly see during pre-training. As a result, attempting to ground the instruction directly on the wrist view is brittle. Instead, we transfer the instruction-aware localization obtained from the base view to the wrist view by treating the object-centric reference crops from previous stage as canonical visual anchors.

Given the wrist-view image \(I^{wrist}\) and its segmentation proposals \(\mathcal{M}^{wrist}=\{m_k^{wrist}\}\), we again employ set-of-mark~\cite{setofmark_2023} and render numeric markers inside every mask, yielding a mark-augmented wrist image. Building on the base-view grounding of previous stage, we reuse the object-centric reference crops associated with each task-relevant object name \(e_j \in \mathcal{E}(l)\). We then form a \textit{single} prompt to the VLM that (i) sequentially lists each object name \(e_j\) together with its reference crop and (ii) appends the mark-augmented wrist-view image. The VLM outputs, for each task-relevant object identified in base view, the marker index corresponding to the same object in the wrist view. These predictions define a subset of wrist-view masks
\begin{equation}
    \mathcal{S}^{wrist} \subseteq \mathcal{M}^{wrist}
\end{equation}

At this point, we obtain instruction-consistent region sets for both cameras,
\(\mathcal{S}^{base} \subseteq \mathcal{M}^{base}\) and \(\mathcal{S}^{wrist} \subseteq \mathcal{M}^{wrist}\), yielding a compact, object-centric description of the scene that is aligned across views.

\textbf{Geometric Grounding.}
Building on the semantic identification of task-relevant objects, the geometric grounding stage constructs representations that capture their underlying 3D structure. We first apply the off-the-shelf Depth Anything v2~\cite{anydepthv2_neurips2024} to RGB images \(I^{base}\) and \(I^{wrist}\) from both views, producing dense depth estimates. To enhance the expressiveness of geometric cues, the grayscale depth values are linearly mapped to a color space with high dynamic range, enabling subtle variations in object structure to be more clearly distinguished. The semantically grounded region sets \(\mathcal{S}^{base}\) and \(\mathcal{S}^{wrist}\) are then applied as masks to depth estimates to filter only the depth measurements associated with relevant objects. The resulting pair of masked depth maps, denoted as \(Z^{base}\) and \(Z^{wrist}\), provide geometry-centered observations that complement the object-centric grounding from previous stage and serve as the comprehensive perceptually grounded visual inputs to the downstream action reasoning module.

\subsection{Perceptually Grounded Action Reasoning via Vision-Language-Action Models}
\label{sec:vla}

As aforementioned
, we employ a VLA model as our policy \(\pi_\theta\), which reasons action from the perceptually grounded visual inputs \(\tilde{o} = (Z^{base}, Z^{wrist})\). This enables the policy to operate over instruction-focused and geometry-aware visual inputs that are substantially less sensitive to visual clutter and appearance variations.

As discussed in \cref{sec:prelim}, we adopt a pretrained VLA as the visuomotor policy \(\pi_\theta\) and fine-tune it on \(\mathcal{D}\) for our robot embodiment. At each time step \(t\), in addition to the perceptually grounded visual inputs \(\tilde{o}_t\), the policy conditions on proprioception \(q_t \in \mathbb{R}^7\), given by the absolute joint angles of the robot and the binary open/close state of the end-effector. It predicts a sequence of future actions \(a_{t:t+H-1}\) for a horizon \(H\), where each element is a 7-dimensional target in the same joint-gripper space.

We only optimize the policy parameters \(\theta\) via the maximum-likelihood objective in \cref{eq:optimization}, while keeping the perceptual grounding module frozen.

\section{Experiments}
\label{sec:experiments}

In this section, we present a suite of experiments to answer the following questions:\\
\textbf{Q1.} Can \model follow fine-grained language instructions in highly distracting scenes?\\
\textbf{Q2.} Can \model remain robust under changes in background appearance and scene layout?\\
\textbf{Q3.} Can \model generalize to manipulating unseen objects in clutter scenes with unseen distractors ?

Additionally, we further conduct ablation studies to probe our perceptual design:\\
\edit{\textbf{Q4.} How much gain is obtained by introducing explicit object-centric grounding before downstream action reasoning?}\\
\textbf{\edit{Q5.}} How crucial is the decoupled two-stage object-centric grounding module in \model?\\
\textbf{\edit{Q6.}} What additional gains does explicit geometry-aware grounding (masked depth inputs) bring over RGB-only grounding?

\subsection{Real-world Setup, Implementation, and Baselines}
\begin{figure}[t]
    \centering
    \includegraphics[width=0.6\linewidth]{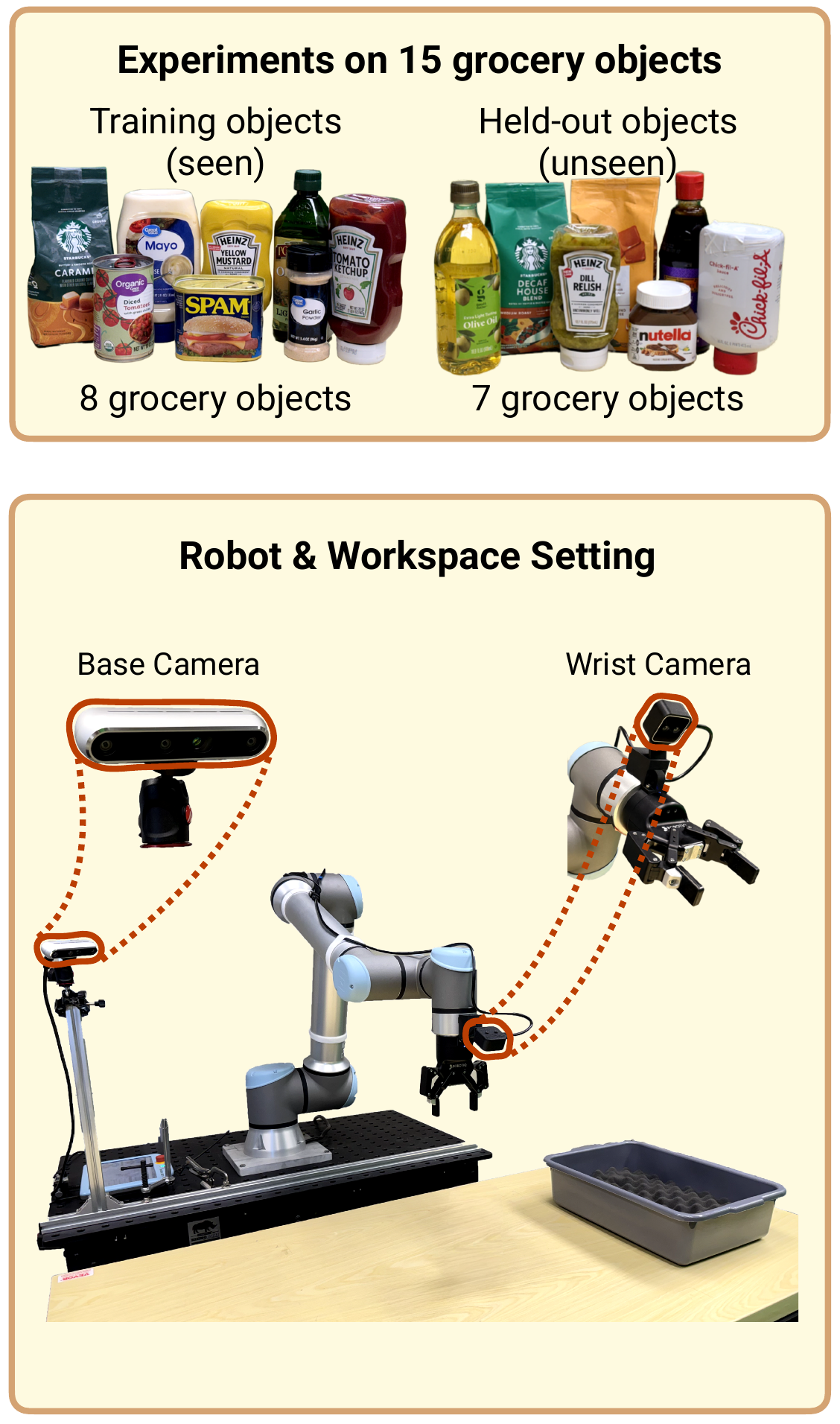}
    \caption{
    \textbf{Experimental~setting:} a UR10e robot with parallel jaw gripper and base/wrist cameras. Policies are trained on single-object pick-and-place demonstrations over eight grocery objects. For evaluation, we test both cluttered scenes built from these training categories and generalization by seven additional object categories that are excluded from training.
    }
    \label{fig:experimental_setup}
\end{figure}

\textbf{Robot platform and control.}
All experiments are conducted on a 6-DoF UR10e robot arm equipped with Robotiq 2F-85 parallel jaw gripper, operating over a tabletop workspace. We capture RGB observations from two synchronized camera streams: a fixed base-view camera placed in an over-the-shoulder viewpoint of the robot, and a wrist-view camera mounted near the wrist--gripper interface. \edit{In our tabletop setup, both RealSense cameras capture \(1280\times720\) RGB streams. We preprocess these RGB observations into \(720\times540\) base- and wrist-view inputs through resizing and center-cropping. As shown in the bottom of \cref{fig:experimental_setup}, the cameras are positioned so that tabletop objects appear at sufficient scale across two views. Demonstrations are collected via teleoperation at a 10\,Hz robot control rate, and the same 10\,Hz rate is used for action execution during deployment.}

\textbf{Training data curation.}
Our training data comprise teleoperated pick-and-place demonstrations collected in \emph{uncluttered} scenes containing a single object placed on the table next to the bin. For each episode, we sample a natural-language instruction from a set of paraphrased templates (listed below) that all specify the same goal of placing the queried object into the bin. The operator then controls the robot to grasp the queried object and place it into the bin. We select eight grocery objects with diverse shapes and appearances, shown in the top of \cref{fig:experimental_setup}, and collect 250 demonstrations per object, resulting in 2000 real-world training demonstrations in total.
The training objects include:
\textit{spice bottle}, \textit{green coffee bag}, \textit{mustard bottle}, \textit{ketchup bottle}, \textit{mayonnaise bottle}, \textit{food can}, \textit{spam tin}, \textit{green oil bottle}.

\textbf{Language prompts.}
We use a small set of paraphrased instruction templates to reduce sensitivity to a single phrasing:
\begin{itemize}\setlength{\itemsep}{0pt}\setlength{\topsep}{2pt}
\item \textit{``place \texttt{<object>} in the bin''}
\item \textit{``put \texttt{<object>} into the bin''}
\item \textit{``pick up \texttt{<object>} and place it in the bin''}
\item \textit{``grasp \texttt{<object>} and drop it into the bin''}
\end{itemize}


\textbf{Implementation details.}

\noindent\emph{Perception grounding module (frozen).} We employ the 8B-Instruct model of Qwen3-VL as the VLM backbone owing to its remarkable reasoning ability at manageable latency. The VLM is deployed using two A6000 GPUs. \edit{For all grounding calls, the VLM receives the preprocessed RGB observations used throughout the system, without additional cropping or upscaling.}

\noindent\emph{Action policy (trainable).} We instantiate the action policy with Pi-0 and Pi-0 FAST backbones. We denote the resulting grounded policies as \textbf{OBEYED Pi-0} and \textbf{OBEYED Pi-0 FAST}. Both models are initialized from the publicly released checkpoints and fine-tuned for 50K iterations with low-rank adaptation~\cite{lora_iclr2022} on the collected demonstrations, using a fixed learning rate of \(1\times 10^{-5}\) and a batch size of 128. Fine-tuning is distributed across four NVIDIA A6000 GPUs, and inference runs on a single GPU on the robot workstation. \edit{Before Pi-0 or Pi-0 FAST inference, the grounded base- and wrist-view observations are resized to \(224\times224\), following the OpenPI policy transform.} Throughout all experiments, the perceptual grounding module is kept frozen; only the VLA policy parameters are updated during fine-tuning. For inference, we follow prior work and execute a cut-off horizon of \(H=10\) actions of the predicted trajectory for both Pi-0 and Pi-0 FAST before querying the policy again from the next observation.

\noindent\edit{\emph{Closed-loop test-time execution.} At test time, the task-aware base-view grounding stage is run only once on the initial frame to identify the instruction-relevant objects, since the scene remains largely static over a rollout and this avoids repeating the expensive base-view grounding step at every control cycle. The selected base-view masks are then propagated across the rest of the rollout. On each control cycle, we run segmentation on the current observations, use the tracked base-view anchors to perform wrist-view cross-view matching, and apply geometric grounding to form the clutter-free observations fed to the VLA. Afterwards, the VLA predicts the next action trajectory, which we execute sequentially. We then obtain the next observation and proceed to the next control cycle.}

\textbf{Baselines.}
We compare \model against current state-of-the-art VLA models, including Pi-0~\cite{pi0_2024}, Pi-0 FAST~\cite{fast_2025}, Pi-0.5~\cite{pi05_2025}, and Gr00T N1.5~\cite{gr00t_2025}. All baselines are initialized from their public checkpoints and fine-tuned on our teleoperated dataset. In contrast to \model, they operate directly on the original RGB observations from two camera views rather than perceptually grounded inputs. For a fair comparison, we use the same optimization hyper-parameters as described above for all baseline models. At test time, we also execute only the first 10 actions from the predicted sequence before re-planning, matching our implementation for \model.

\begin{figure*}[t]
    \centering
    \includegraphics[width=0.95\linewidth]{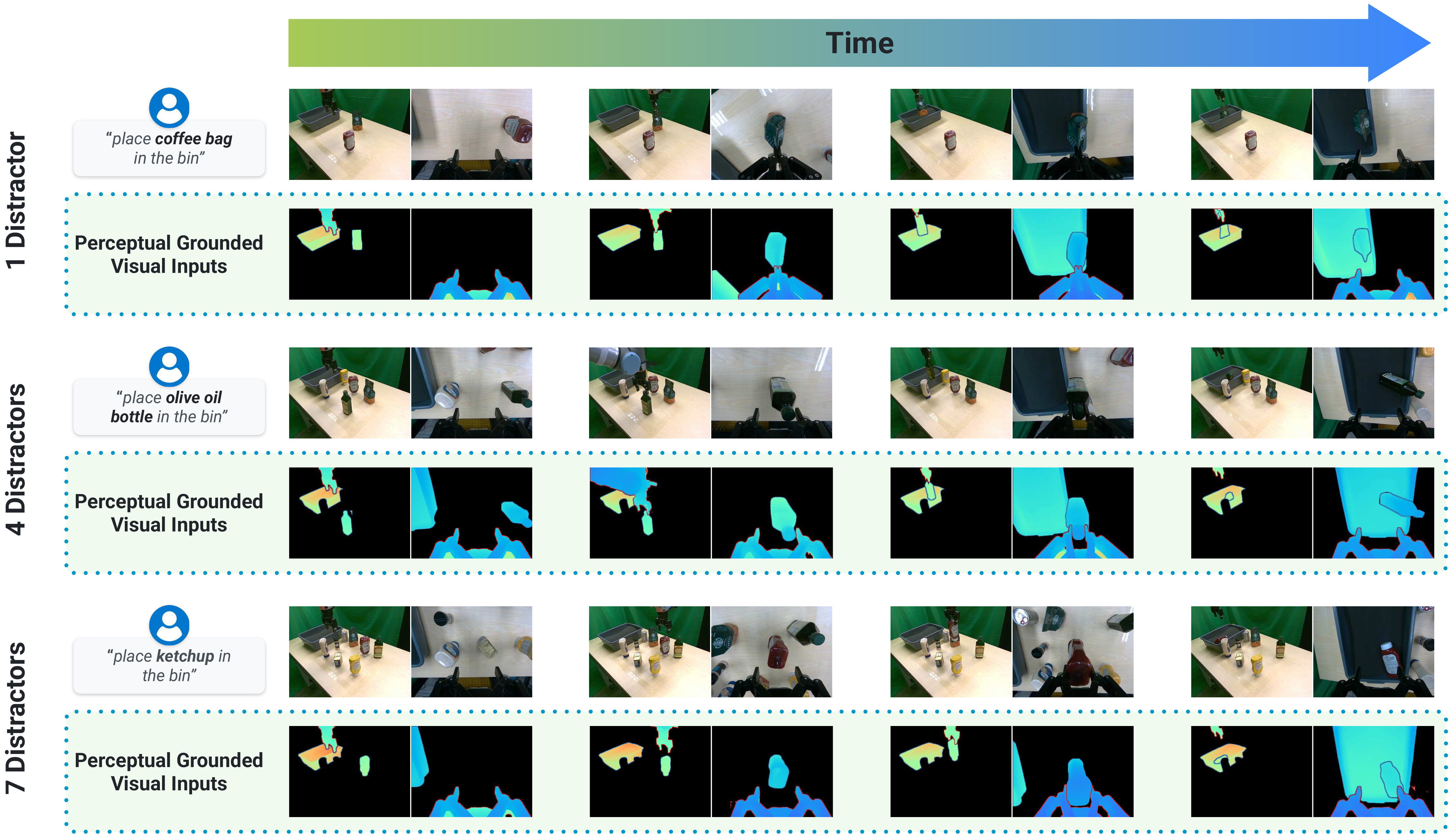}
    \caption{\textbf{Qualitative trials in cluttered scenes with distractors sampled from eight training objects.} For each instruction, we show the original RGB observations and the corresponding perception-grounded views produced by \modelPi. The grounded inputs suppress distractor objects and highlight the queried target, allowing the policy to ignore clutter and precisely execute the task.}
    \label{fig:qualitative_distr}
\end{figure*}

\subsection{Fine-grained language following in distracting scenes}
\label{sec:lang_following}

\textbf{Experimental setting.}
To answer \textbf{Q1}, we introduce three real-world tabletop benchmarks that stress fine-grained, language-conditioned grounding in the presence of visual distractors. Each trial consists of a natural-language instruction and a tabletop scene; the policy must either execute the correct pick-and-place on the queried object or refrain from acting when the instruction is inconsistent with the scene.

\emph{(1) Distractor objects.}
We populate the workspace with one target object and multiple distractors, all sampled from the eight training categories. Distractors are objects that are not referenced by the current instruction. The instruction names exactly one visible object, and success requires picking only that target.

\emph{(2) Absent-target rejection.}
We place a single object on the table but issue an instruction that refers to a different object category that is not present. The correct behavior is to reject the instruction by not picking any object. This setting explicitly probes a model’s tendency to over-act on spurious visual cues rather than enforcing consistency between language and scene.

\emph{(3) Spatial reasoning.}
We uniformly sample three objects, place them in a horizontal row with randomized ordering, and issue purely relational instructions (e.g., \textit{``place the left object in the bin''}). This task forces the policy to rely on relational and spatial reasoning rather than category- or appearance-based matching.

\textbf{Evaluation protocol.}
For the distractor object task, we evaluate three difficulty levels with \{1, 4, 7\} distractors around a single target, randomly sampling objects and placements in each trial. For the absent-target task, we rollout trials by pairing a physically present object with an instruction that names a different object, and count success only when no pick is executed. For relational grounding, each trial samples three objects and randomly permutes their positions (left, middle, right) on the table. Across all tasks and difficulty levels, we report success rate and confidence interval (CI) over 100 rollouts per model and configuration. Together, these benchmarks expose complementary failure modes: confusion under heavy clutter, over-confident grasping on infeasible instructions, and weak generalization to relational language grounding.

\begin{figure}[t]
    \centering
    \includegraphics[width=\linewidth]{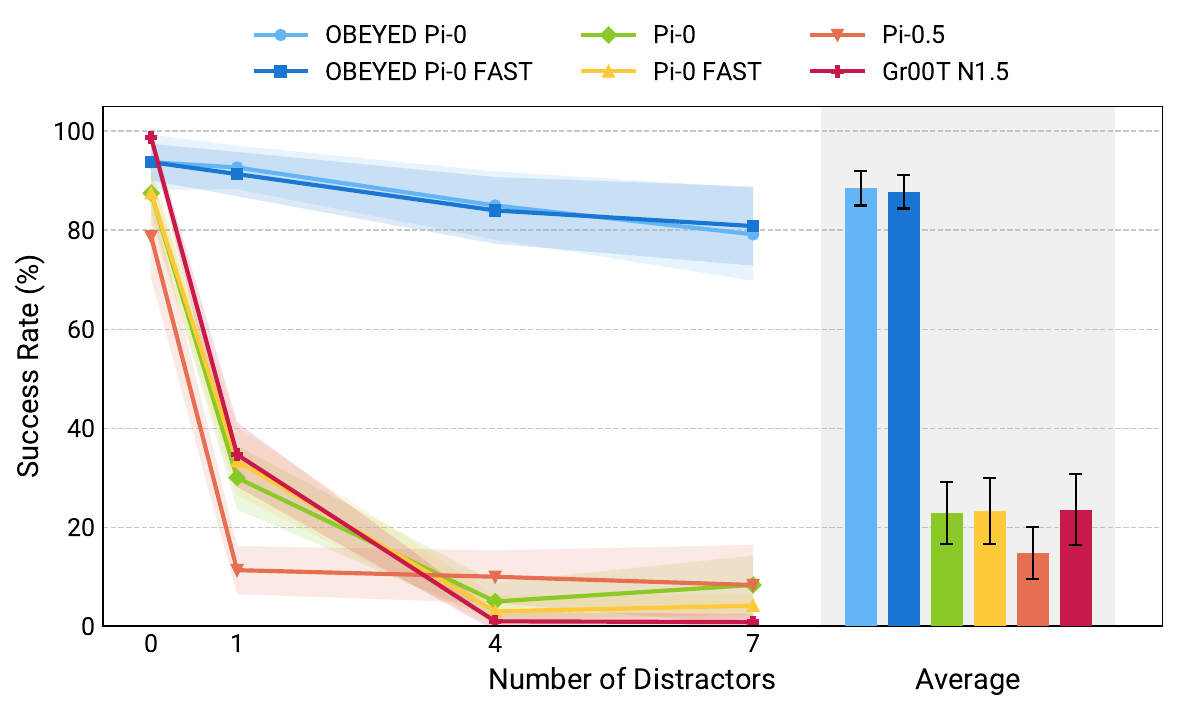}
    \caption{\textbf{Success rate (\%) on fine-grained language following with distractors sampled from eight training objects.} Comparison between \model and state-of-the-art VLAs as we increase the number of distractors from 0 (distractor-free) to 1, 4, and 7. We report mean success with 95\% CI.}
    \label{fig:exp_distractor}
\end{figure}

\begin{figure}[t]
    \centering
    \includegraphics[width=\linewidth]{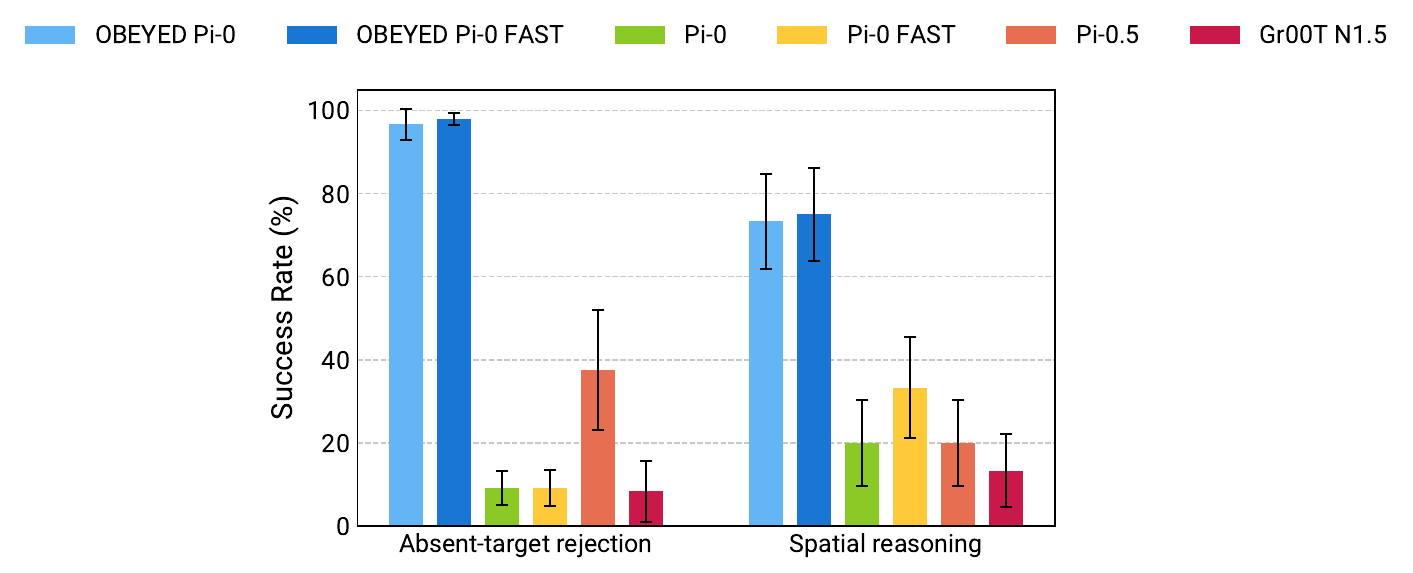}
    \caption{\textbf{Success rate (\%) on absent-target rejection and spatial reasoning benchmarks.} Absent-target rejection measures how often a policy correctly refrains from grasping when the requested object is missing, while spatial reasoning evaluates following spatially relational instructions (e.g., ``left object''). We report mean success with 95\% CI.}
    \label{fig:exp_absent_spatial}
\end{figure}

\textbf{Results and analysis.}
\cref{fig:exp_distractor} compares \modelPi and \modelFAST against state-of-the-art VLAs as the number of distractors increases. In the distractor-free setting (0 distractors), all methods achieve high success rates (\(\geq 80\%\)). However, as we add more distractor objects, prior VLAs drastically collapse to below \(10\%\), whereas both \modelPi and \modelFAST remain above \(90\%\) with one distractor and around \(80\%\) even in the heaviest clutter regime. Averaged across all clutter levels, both instances of our framework yield \(4\times\) improvement over the strongest baseline, indicating that our perception-grounded design largely prevents clutter-induced collapse and enables reliable fine-grained language following in densely populated scenes. Qualitative rollouts in \cref{fig:qualitative_distr} further show that across all difficulty levels, \modelPi consistently grounds the instruction on the correct relevant objects, maintains attention on these objects throughout the approach and grasp, and ignores nearby distractors even when they are visually closer to the gripper.

We summarize results on absent-target rejection and spatial reasoning tasks in \cref{fig:exp_absent_spatial}. For absent-target rejection, both \modelPi and \modelFAST achieve nearly perfect success (\(\sim 95\%\)), whereas Pi-0.5 reaches at most \(\sim 40\%\) and the remaining VLAs stay around \(10-15\%\), revealing a strong tendency to execute spurious grasps when no valid target is present. For spatial reasoning, where category cues are uninformative and the policy must rely purely on spatial information, \model attains \(\sim 75\%\) success on both instances, outperforming the best baseline (Pi-0 FAST) by over \(40\) absolute points. These results show that our perception-grounded framework substantially strengthens feasibility checking and relational grounding beyond what current end-to-end VLAs exhibit.

Supplementary video illustrates representative rollouts across distractor settings (0–7 distractors) as well as the absent-target rejection and spatial reasoning tasks.

\begin{figure}[t]
    \centering
    \includegraphics[width=\linewidth]{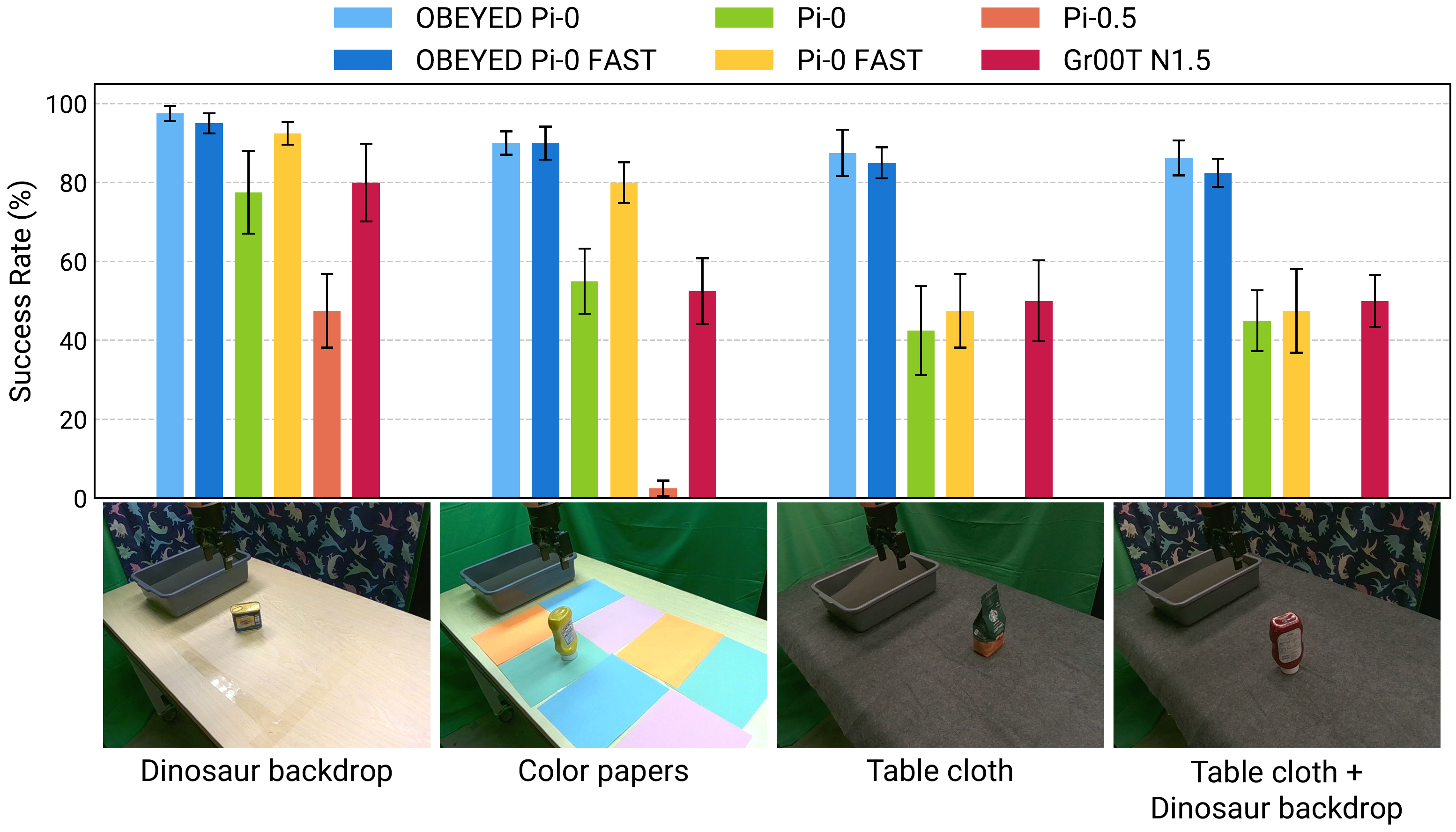}
    \caption{\textbf{Success rate (\%) on out-of-distribution background shifts.}
    We quantitatively compare \model and state-of-the-art VLAs across four background variants, from mild to severe table and backdrop changes. We report mean success with 95\% CI.
    }
    \label{fig:exp_bg_changes}
\end{figure}

\subsection{Robustness to background changes}

\textbf{Experimental setting.}
In addressing \textbf{Q2}, we specifically test how background appearance affects policy performance under the simplest interaction setting: a single target object on the table and a language instruction that exactly matches that object. This isolates the effect of background changes from clutter and instruction ambiguity. Prior work~\cite{byovla_icra2025} has shown that VLAs can be brittle to background shifts; here we examine whether our perceptual grounding module improves robustness in such cases.

\begin{figure*}[t]
    \centering
    \includegraphics[width=0.95\linewidth]{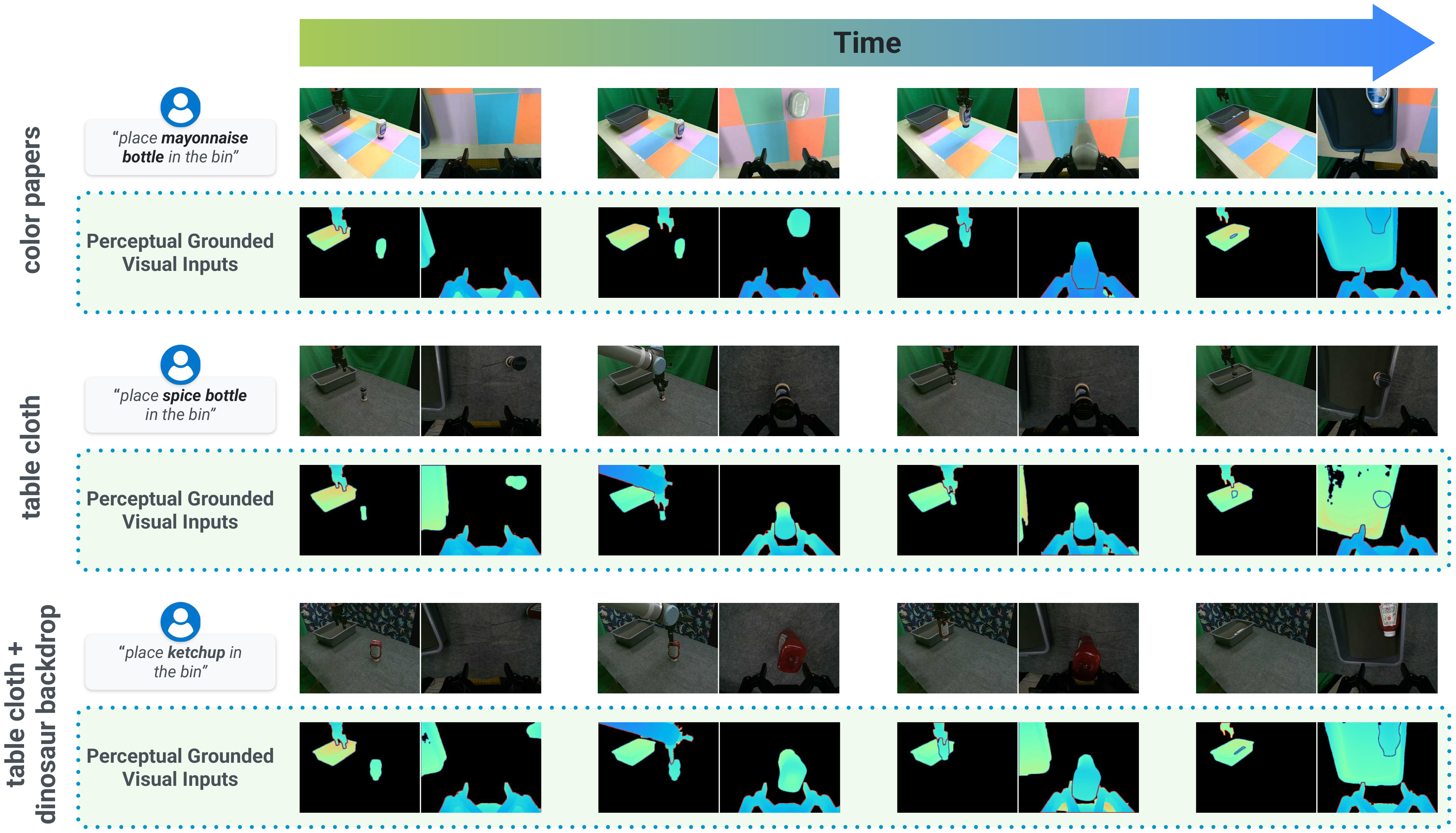}
    \caption{\textbf{Qualitative results under background appearance shifts.} Example rollouts under different out-of-distribution backgrounds, showing the original RGB observations and the corresponding perception-grounded views. The grounded inputs suppress distracting background variation around the target object and receptacle, enabling the policy to consistently execute the given task despite large changes in surrounding appearance.}
    \label{fig:qualitative_bg}
\end{figure*}

\textbf{Evaluation protocol.}
We evaluate four background variants, ordered by increasing background shift severity: (1) placing a tablecloth with a completely different color and pattern on the tabletop, (2) replacing the backdrop with a different visual scene, (3) randomly spreading multi-colored papers on the table, and (4) combining both the new tablecloth and new backdrop. For each background condition, we run 50 rollouts per model and report success rate with 95\% CI.

\textbf{Results and analysis.}
\cref{fig:exp_bg_changes} reports success rates under four out-of-distribution backgrounds. Across all conditions, \modelPi remains highly stable (\(\geq 80\%\)) with only modest degradation from the clutter-free single object setting, whereas all baselines exhibit substantial drops. The dinosaur backdrop alone causes only mild degradation for Pi-0, Pi-0 FAST, and Gr00T N1.5, but drives a sharp decline for Pi-0.5, indicating its poor generalization. The largest performance losses occur when perturbations affect regions in direct contact with the object: color papers and the tablecloth typically reduce baseline success by roughly \(10\text{--}15\) and an additional \(5\text{--}15\) absolute points, respectively, with Pi-0.5 collapsing to near-zero under color papers. Adding the dinosaur backdrop on top of the tablecloth produces unnoticeable changes, suggesting that shifts in the tabletop region dominate over distant background changes. In contrast, \modelPi degrades only slightly across this spectrum of background shifts, highlighting that our explicit object-centric grounding substantially mitigates background overfitting. In all background settings, \modelPi and \modelFAST achieve comparable success rates, suggesting that our framework is largely policy-agnostic and consistently improves robustness to background-induced distractors. In \cref{fig:qualitative_bg}, the perception-grounded views remain visually consistent across color papers, tablecloth, and tablecloth+backdrop backgrounds, while suppressing large appearance variation in the raw RGB observations. As a result, the policy maintains stable focus on the instruction-relevant target and receptacle throughout the rollout despite substantial background shifts.

Supplementary video illustrates representative rollouts across all background shift scenarios.

\subsection{Fine-grained language following on unseen objects}
\textbf{Experimental setting.}
To address \textbf{Q3}, we evaluate whether the policies can correctly perceive language instructions that name novel objects and act in scenes composed entirely of unseen objects. We construct the \textit{distractor objects} task with seven held-out grocery objects disjoint from the eight training objects, as shown in \cref{fig:experimental_setup}. In the scene, all objects are randomly placed on the table. The instruction follows the same language-following format as before but now names a single unseen category, and the policy must complete the pick-and-place on the queried unseen object while ignoring unseen distractors.
The list of unseen objects (shown in \cref{fig:experimental_setup}) includes:
\textit{green coffee bag}, \textit{orange coffee bag}, \textit{white sauce bottle}, \textit{hoisin sauce bottle}, \textit{relish bottle}, \textit{nutella}, \textit{yellow oil bottle}.

\textbf{Evaluation protocol.}
We adopt the same evaluation protocol as discussed in \textit{distractor objects} task. We run 100 rollouts per model and report the success rate with 95\% CI.

\begin{figure}
    \centering
    \includegraphics[width=0.5\linewidth]{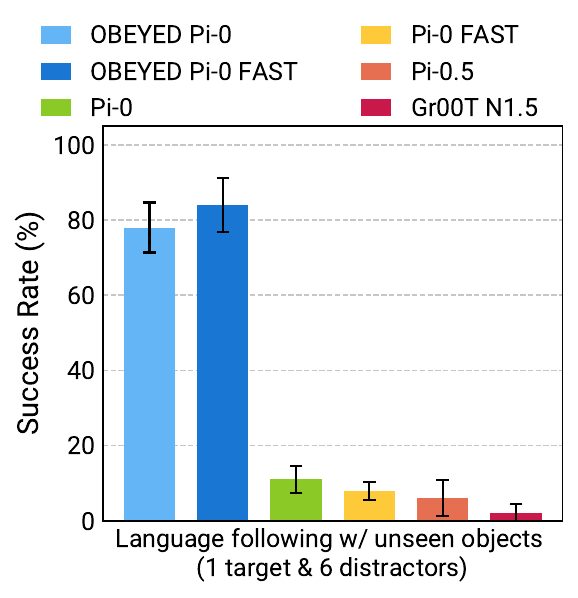}
    \caption{\textbf{Success rate (\%) on fine-grained language following with unseen objects under clutter.}
    Each scene contains one unseen target object and four unseen distractors sampled from seven held-out categories, and the instruction names the unseen target category. We report mean success with 95\% CI.}
    \label{fig:exp_unseen_obj}
\end{figure}

\textbf{Results and analysis.}
\cref{fig:exp_unseen_obj} shows that \modelFAST consistently achieves the highest success rate on unseen-object language following, even though every object in the scene belongs to a novel category. Similar to the \emph{seen distractor objects} setting, standard Pi-0 and Pi-0 FAST suffer substantial drops and Pi-0.5 and Gr00T N1.5 nearly fail under unseen clutter, whereas \modelFAST sustains high performance and remains far above all baselines. These results confirm that explicit object-centric and geometry-aware grounding is key to enable reliable transfer of visuomotor skills to novel objects in realistic, cluttered scenes. On this unseen setting, \modelPi closely tracks \modelFAST, trailing by only \(\sim 5\%\) absolute success.
The supplementary video further showcases representative rollouts in these unseen-object clutter scenes, highlighting the ability of \model to follow language instructions even on novel objects.

\subsection{Ablation Studies}

\edit{\textbf{Effect of explicit object-centric grounding (Q4).}}
\edit{To measure the gain from introducing a decoupled perception-control interface, we compare baseline Pi-0 (5$^{\textnormal{th}}$ row of \cref{tab:ablation}) against the configuration with single-stage object-centric grounding only (4$^{\textnormal{th}}$ row). In this setting, grounding is applied independently to the current base and wrist images at each control cycle, without cross-view region matching or geometric grounding, and the resulting masked RGB observations are fed directly to Pi-0. Even this minimal perception grounding already produces large gains across all three settings, especially on cluttered seen-object and unseen-object manipulation. This result indicates that a major portion of the robustness gain begins with explicitly filtering the scene to task-relevant objects before downstream action reasoning.}

\textbf{Effect of two-stage object-centric grounding (\edit{Q5}).}
\edit{To quantify the importance of our two-stage object-centric grounding, we compare the full model (1$^{\textnormal{st}}$ row of \cref{tab:ablation}) against the single-stage counterpart with geometry-aware grounding enabled (2$^{\textnormal{nd}}$ row), and the RGB-only two-stage configuration (3$^{\textnormal{rd}}$ row) against the RGB-only single-stage configuration (4$^{\textnormal{th}}$ row). In both comparisons, the single-stage version grounds the base and wrist views independently rather than using base-view references to guide wrist-view matching. This replacement causes clear degradation across the benchmarks, with the largest drop appearing on spatial reasoning and additional reductions on both seen-object and unseen-object clutter. These consistent gaps show that base-view reference crops are crucial for resolving wrist-view ambiguity under partial visibility and relational prompts, where single-stage prompting tends to lock onto visually salient but incorrect regions.}

\textbf{Effect of geometry-aware grounding (\edit{Q6}).}
\edit{To isolate the contribution of explicit geometry-aware grounding, we compare the full model (1$^{\textnormal{st}}$ row of \cref{tab:ablation}) against the RGB-only two-stage configuration (3$^{\textnormal{rd}}$ row), and the single-stage geometry-aware configuration (2$^{\textnormal{nd}}$ row) against the single-stage RGB-only configuration (4$^{\textnormal{th}}$ row). In these RGB-only settings, the downstream VLA receives masked RGB observations instead of masked depth-based inputs. The clearest effect of geometry-aware grounding appears on unseen-object clutter, where adding depth cues consistently improves performance in both pairwise comparisons. By contrast, the gains on seen-object clutter and spatial reasoning are smaller. This pattern indicates that masked depth cues provide complementary geometric structure beyond object-centric filtering alone, reducing the reliance of action reasoning on texture and color.}

\begin{table}[!t]
\caption{Ablation of object-centric and geometric grounding in \model, with Pi-0~\cite{pi0_2024} as the downstream action policy.
}
\label{tab:ablation}
\centering
\footnotesize
\setlength{\tabcolsep}{2pt}
\resizebox{\columnwidth}{!}{%
\begin{tabular}{ccc|ccc}
\toprule
\multicolumn{3}{c|}{\edit{\textbf{Modules}}} & \multicolumn{2}{c}{\textit{\textbf{Seen targets}}} & \textit{\textbf{Unseen targets}} \\
\edit{OC} & \edit{2-stage OC} & \edit{Geometric} & 4 Distr. & Spatial & 4 Distr. \\
\edit{grounding} & \edit{grounding} & \edit{grounding} & (seen obj.) & Reasoning & (unseen obj.) \\
\midrule
\cmark & \cmark & \cmark & $85 \pm 6.9$ & $73 \pm 11.4$ & $78 \pm 6.7$ \\
\cmark & \xmark & \cmark & $69 \pm 2.4$ & $43 \pm 2.9$ & $69 \pm 2.7$ \\
\cmark & \cmark & \xmark & $83 \pm 2.0$ & $70 \pm 2.5$ & $70 \pm 2.6$ \\
\edit{\cmark} & \edit{\xmark} & \edit{\xmark} & \edit{$71 \pm 5.4$} & \edit{$37 \pm 3.8$} & \edit{$57 \pm 3.3$} \\
\edit{\xmark} & \edit{\xmark} & \edit{\xmark} & \edit{$5 \pm 3.9$} & \edit{$20 \pm 10.3$} & \edit{$11 \pm 3.7$} \\
\bottomrule
\end{tabular}%
}\vspace{0.3em}
{\edit{\footnotesize OC grounding = Object-centric grounding}}
\end{table}

\subsection{Run-time analysis}
\label{sec:runtime_analysis}

Inference time of \model is decomposed into four components: (1) the segmentation proposal module, which predicts object masks for both views; (2) the object-centric grounding stage, which includes cross-view region matching and invokes the VLM to associate wrist-view crops with task-relevant base-view objects; (3) the geometric grounding stage, which back-projects selected masks into 3D and forms object-centric crops; and (4) the VLA policy (Pi-0 or Pi-0 FAST), which decodes the next action sequence. We profile these components on the robot workstation and report their per-step wall-clock latency in \cref{tab:runtime}, averaging over 10 rollouts. Segmentation and grounding costs are shared across all variants, while the policy time reflects the specific backbone.

Object-centric grounding runs at \(0.41\,\mathrm{s}\) per inference call on average for cross-view region matching. Given that the scene remains largely static over a rollout, we perform task-aware grounding \edit{on base-view observation} only once at initialization frame and subsequently rely on segmentation model to propagate the selected masks across frames. The remaining components--Segmentation (\(0.04\,\mathrm{s}\)), geometric grounding (\(0.18\,\mathrm{s}\)), and action policy inference (\(0.15\,\mathrm{s}\) for Pi-0, \(0.53\,\mathrm{s}\) for Pi-0 FAST)--also operate within the sub-second range. Overall, this yields end-to-end control cycles of \edit{\(0.78\,\mathrm{s}\)} with Pi-0 and \(1.16\,\mathrm{s}\) with Pi-0 FAST \edit{in the sequential setting} (around \edit{\(0.8\text{--}1.2\,\mathrm{Hz}\)}), which is sufficient for our real-world tabletop manipulation tasks.

\edit{\textbf{Geometric grounding in parallel.} A straightforward approach to improve inference speed is to parallelize independent modules. In our pipeline, the depth-estimation step in geometric grounding operates directly on the incoming RGB observations, so we run this branch in a separate process alongside segmentation and object-centric grounding; once the selected masks are available, only lightweight mask composition remains before VLA inference. We implement this parallel execution and report the measured end-to-end latencies in the lower comparison rows of \cref{tab:runtime}.

Compared with the sequential setting, this reduces the single-step latency by about \(21\%\) (\(0.62\,\mathrm{s}\) vs. \(0.78\,\mathrm{s}\)) for OBEYED Pi-0 and \(15\%\) (\(0.99\,\mathrm{s}\) vs. \(1.16\,\mathrm{s}\)) for OBEYED Pi-0 FAST.}

\begin{table}[t]
  \centering
  \caption{Run-time of \model for a single inference pass, averaged over 10 rollouts.}
  \setlength{\tabcolsep}{2pt}
  \label{tab:runtime}
  \begin{tabular}{lc}
    \toprule
    \textbf{Stage}         & \textbf{Single inference call (s)} \\
    \midrule
    Segmentation proposals & \textit{0.04} \\
    Object-centric grounding     & \textit{0.41} \\
    Geometric grounding    & \textit{0.18} \\
    \midrule
    VLA policy (Pi-0)      & \textit{0.15} \\
    VLA policy (Pi-0 FAST) & \textit{0.53} \\
    \midrule
    \textbf{OBEYED Pi-0}            &  \\
    \edit{\hspace{0.2em}\raisebox{0.25ex}{$\llcorner$} sequential} & \edit{\textit{0.78}} \\
    \edit{\hspace{0.2em}\raisebox{0.25ex}{$\llcorner$} with geometric grounding in parallel} & \edit{\textit{0.62}} \\
    \textbf{OBEYED Pi-0 FAST}       &  \\
    \edit{\hspace{0.2em}\raisebox{0.25ex}{$\llcorner$} sequential} & \textit{1.16} \\
    \edit{\hspace{0.2em}\raisebox{0.25ex}{$\llcorner$} with geometric grounding in parallel} & \edit{\textit{0.99}} \\
    \bottomrule
  \end{tabular}
\end{table}

\begin{table}[t]
  \centering
  \caption{\textcolor{RevisionColor}{Average total rollout time over 10 rollouts under different runtime optimizations.}}
  \setlength{\tabcolsep}{2pt}
  \label{tab:runtime_rollout}
  \footnotesize
  \begin{tabular}{lc}
    \toprule
    \textcolor{RevisionColor}{\textbf{Variant}} & \textcolor{RevisionColor}{\textbf{Avg. rollout time (s)}} \\
    \midrule
    \textcolor{RevisionColor}{\textbf{OBEYED Pi-0}} &  \\
    \textcolor{RevisionColor}{\hspace{0.2em}\raisebox{0.25ex}{$\llcorner$} sequential} & \textcolor{RevisionColor}{\textit{34.26}} \\
    \textcolor{RevisionColor}{\hspace{0.2em}\raisebox{0.25ex}{$\llcorner$} with geometric grounding in parallel} & \textcolor{RevisionColor}{\textit{32.91}} \\
    \textcolor{RevisionColor}{\hspace{0.2em}\raisebox{0.25ex}{$\llcorner$} with object-centric grounding gating} & \textcolor{RevisionColor}{\textit{30.70}} \\
    \textcolor{RevisionColor}{\hspace{0.2em}\raisebox{0.25ex}{$\llcorner$} with both} & \textcolor{RevisionColor}{\textit{29.83}} \\
    \textcolor{RevisionColor}{\textbf{Baseline Pi-0}} & \textcolor{RevisionColor}{\textit{26.45}} \\
    \midrule
    \textcolor{RevisionColor}{\textbf{OBEYED Pi-0 FAST}} &  \\
    \textcolor{RevisionColor}{\hspace{0.2em}\raisebox{0.25ex}{$\llcorner$} sequential} & \textcolor{RevisionColor}{\textit{43.77}} \\
    \textcolor{RevisionColor}{\hspace{0.2em}\raisebox{0.25ex}{$\llcorner$} with geometric grounding in parallel} & \textcolor{RevisionColor}{\textit{40.19}} \\
    \textcolor{RevisionColor}{\hspace{0.2em}\raisebox{0.25ex}{$\llcorner$} with object-centric grounding gating} & \textcolor{RevisionColor}{\textit{35.28}} \\
    \textcolor{RevisionColor}{\hspace{0.2em}\raisebox{0.25ex}{$\llcorner$} with both} & \textcolor{RevisionColor}{\textit{32.89}} \\
    \textcolor{RevisionColor}{\textbf{Baseline Pi-0 FAST}} & \textcolor{RevisionColor}{\textit{28.96}} \\
    \bottomrule
  \end{tabular}
\end{table}

\edit{\textbf{Object-centric grounding gating.} Another way to reduce inference cost is to invoke object-centric grounding only when the scene content changes, rather than at every control cycle. Concretely, after object-centric grounding identifies the task-relevant objects in the current frame, we employ ByteTrack~\cite{bytetrack_eccv2022} to track these selected objects across subsequent frames. We reuse the previous grounding result as long as no new object is detected and no tracked task-relevant object disappears; otherwise, we trigger a new grounding call. This gating mechanism reduces unnecessary VLM calls while preserving the same grounded-input interface to the downstream VLA. We report the resulting average rollout times over 10 rollouts in \cref{tab:runtime_rollout}. Unlike \cref{tab:runtime}, this table measures end-to-end wall-clock rollout time, which includes not only model inference but also physical robot execution, during which the control loop is idle.

This strategy is particularly effective over long rollouts, where many consecutive control cycles share the same tracked object set. Skipping unnecessary object-centric grounding calls in those intervals yields a larger rollout-time reduction than parallel geometric grounding alone. When combined with parallel geometric grounding, the grounded system operates within about \(13\%\) of the raw Pi-0 baseline (\(29.83\,\mathrm{s}\) vs. \(26.45\,\mathrm{s}\)) and about \(14\%\) of the raw Pi-0 FAST baseline (\(32.89\,\mathrm{s}\) vs. \(28.96\,\mathrm{s}\)). This is achieved while retaining the robustness gains shown in the main experiments.}

\section{Conclusion}
\textbf{Summary of contributions.}
We introduced \model, an object-centric \& geometry grounded vision--language--action framework that explicitly decouples visual grounding from action reasoning. Rather than relying on a monolithic end-to-end VLA model, \model augments an arbitrary VLA with a modular, frozen perception pipeline that produces task-conditioned, object-centric, and geometry-aware observations from raw multi-view RGB inputs. Concretely, a VLM-driven object-centric grounding module identifies instruction-relevant regions across multiple camera views via set-of-mark prompting, while a geometric grounding module converts these regions into masked depth representations that concentrate on 3D structure over appearance. The resulting perceptually grounded visual inputs are then fed to a pretrained VLA policy, which is fine-tuned only on clean, single-object demonstrations while the grounding modules remain frozen.

On a real-world UR10e tabletop setup, we validated \model across four challenging deployment regimes—(i) clutter with distractor objects, (ii) absent-target instruction rejection, (iii) background appearance shifts, and (iv) cluttered manipulation of unseen objects\edit{, corresponding to our six experimental questions (Q1--Q6) on robustness, generalization, and the roles of object-centric grounding, two-stage object-centric grounding, and explicit geometry-aware grounding}. Across these settings, \model consistently improves reliability and generalization over strong VLA baselines without requiring synthetic clutter generation or auxiliary perceptual training objectives during VLA fine-tuning. Ablations further confirm that \edit{explicit object-centric grounding provides the main perception-control interface, while} two-stage object-centric grounding and geometry-aware grounding \edit{contribute additional} complementary gains. Overall, our results suggest that treating perception grounding as an explicit, modular component is an effective and complementary path to making VLA policies more reliable in clutter, more focused under distractors, and more transferable to unseen objects and backgrounds.

\textbf{Limitations and future directions.}
Our framework also motivates several future extensions. First, \model depends on the reliability of its perception components (segmentation, VLM-based grounding, and depth estimation). For example, if the segmentation network merges nearby instances in dense clutter, the resulting grounded views can become imperfect and may reduce downstream action accuracy. \edit{In our setting, such failures mainly arise at inference time under very dense clutter, heavy occlusion, or tight object spacing. On the clean single-object demonstrations used for training, our perception grounding module is highly reliable, and we do not observe noticeable mask errors. These severe-occlusion cases most often affect the base view, where the target can be partially hidden behind nearby objects and the grounded region may include both the target and occluder. Even when the wrist view provides a cleaner observation, the downstream VLA still receives the imperfect base-view input, which can degrade action prediction. A promising direction is to make the downstream generative policy more tolerant to imperfect grounded observations through inference-time steering or guidance from auxiliary reward or dynamics models.}

\edit{Second, in this paper we focus on improving robustness under cluttered scenes, while efficiency is not our primary focus. We nevertheless apply two simple runtime optimizations, namely running depth estimation in a separate process to enable parallel execution, and gating wrist-view object-centric grounding so it is invoked only when the tracked scene content changes. Even with these improvements,} the use of off-the-shelf modules for segmentation, VLM inference, and depth estimation introduces non-trivial \edit{computational overhead}. In settings where an RGB-D sensor is available, depth can be obtained directly from the camera, eliminating the external depth estimator and reducing overhead. More broadly, a promising direction is to distill or amortize the grounding pipeline into lighter models, or to equip and train VLAs with an internal object-centric grounding stage following our explicit pipeline as supervision.

Finally, since our goal in this work is to establish the effectiveness of explicit perception grounding, our experiments have been conducted on short-horizon tabletop pick-and-place; extending our framework to long-horizon, multi-stage tasks and more dynamic environments remains an important direction. \edit{In addition, depth is used here to emphasize object geometry for grounding and generalization. However, geometric information can also support collision avoidance and trajectory planning, so extending the grounded representation toward obstacle-aware action reasoning is a natural future direction.}


\bibliographystyle{IEEEtran}
\bibliography{references}

\clearpage
\appendices

\section{\edit{YOLO11-Seg Training Details}}
\label{app:yolo_training}

\edit{For reproducibility, we detail here additional training details of the YOLO11-Seg model used in \cref{sec:method_pg} for object mask proposal generation. Our training data are constructed from two sources.

First, we use 100 teleoperated robot demonstrations collected in our tabletop setup, with 80 demonstrations for training and 20 for validation, and automatically annotate them with the same tools described in the main text: Co-DETR provides whole-object masks for workspace items, while Grounding DINO, SAM, and Cutie are used to localize and propagate the robot arm and gripper masks across frames. After dataset preparation, this robot demo source contains 76,871 training images and 8,818 validation images. These images expose the segmentation model to the camera viewpoints, manipulator appearance, and scene layout encountered at deployment time.}

\edit{Second, to ensure that the segmentation model generalizes beyond the limited set of seen grocery objects in our robot demonstrations, we add a filtered LVIS v1 subset containing indoor tabletop categories. We tailor a list of indoor object class names and remove a few ambiguous classes so that the resulting LVIS subset remains approximately matched in scale to the robot demo source. We then retain LVIS images that contain at least one selected category, resulting in 70,813 training images and 13,732 validation images. This allows the robot demo data and the filtered LVIS subset to be sampled in an approximately 50:50 ratio during fine-tuning.

During training, we apply image augmentations including MixUp image interpolation with probability 0.15 and Copy-Paste object insertion with probability 0.30. We additionally use random horizontal flipping with probability 0.5, small random rotations up to \(\pm 3^\circ\), random rescaling within \(\pm 15\%\), and shear augmentation with magnitude 0.5. \Cref{tab:yolo_training} summarizes the main fine-tuning parameters.}

\begin{table}[h]
  \centering
  \caption{\textcolor{RevisionColor}{Summary of YOLO11-Seg fine-tuning details used in Section~IV.A.}}
  \label{tab:yolo_training}
  \footnotesize
  {\color{RevisionColor}
  \begin{tabular}{@{}p{2.8cm}p{4.8cm}@{}}
    \toprule
    \textbf{Item} & \textbf{Setting} \\
    \midrule
    Model initialization & \texttt{yolo11l-seg} pretrained checkpoint \\
    Input resolution & Longer side resized to 960 \\
    Batch size & 16 \\
    Optimizer & AdamW \\
    Training schedule & 5 epochs \\
    Devices & 2 A6000 GPUs (48GB each)  \\
    Training time & \(\sim 12\) hours \\
    \bottomrule
  \end{tabular}
  }
\end{table}

\section{\edit{Additional Experiments and Failure Analysis}}
\label{app:additional_experiments}

\subsection{\edit{Prediction Horizon Sensitivity}}
\edit{In the main paper, OBEYED Pi-0 executes a cut-off horizon of \(H=10\) actions before querying the policy again from the next observation. Here, we study the sensitivity of our framework to the choice of prediction horizon \(H\). As a representative clutter setting, we evaluate the case of one target with four distractors, where all objects belong to the seen training categories. We vary the execution horizon over \(H\in\{5,10,15,20\}\) while keeping the rest of the deployment pipeline unchanged. Results are summarized in \Cref{tab:horizon_sensitivity}.}

\begin{table}[h]
  \centering
  \caption{\edit{Prediction horizon sensitivity on the representative clutter setting of one target with four seen distractors.}}
  \label{tab:horizon_sensitivity}
  \footnotesize
  \setlength{\tabcolsep}{6pt}
  \renewcommand{\arraystretch}{1.1}
  \begin{tabular}{lcc}
    \toprule
    \edit{\textbf{Action Horizon \(H\)}} & \edit{\textbf{Pick-up}} & \edit{\textbf{Pick-and-Place}} \\
     & \edit{\textbf{Success Rate}} & \edit{\textbf{Success Rate}} \\
    \midrule
    \edit{\(5\)}  & \edit{\textit{88}} & \edit{\textit{81}} \\
    \edit{\(10\) (default)} & \edit{\textit{87}} & \edit{\textit{85}} \\
    \edit{\(15\)} & \edit{\textit{77}} & \edit{\textit{76}} \\
    \edit{\(20\)} & \edit{\textit{56}} & \edit{\textit{56}} \\
    \bottomrule
  \end{tabular}
\end{table}

\edit{Overall, varying the action horizon mainly affects control execution rather than perception grounding: across all tested horizons, the framework still grounds the target correctly and approaches it successfully in most rollouts. A shorter horizon (\(H=5\)) gives the highest pick-up success, suggesting fewer overshooting failures (i.e., the robot hits the target object while approaching it), but its final pick-and-place success drops below the default \(H=10\), indicating additional failures during the release phase. In particular, after reaching the bin, the robot often moves back and forth above it without releasing the object, suggesting that the releasing motion itself requires longer than five actions and that the cut-off horizon truncates the gripper-opening phase. In contrast, longer horizons (\(H\geq 15\)) reduce both pick-up and final success, consistent with more overshooting and object-contact failures during approach. Therefore, among the tested settings, \(H=10\) yields the most reliable overall behavior in our setup, which also supports our choice to follow prior work in using this default horizon.}

\begin{figure*}[t]
    \centering
    \includegraphics[width=0.95\textwidth]{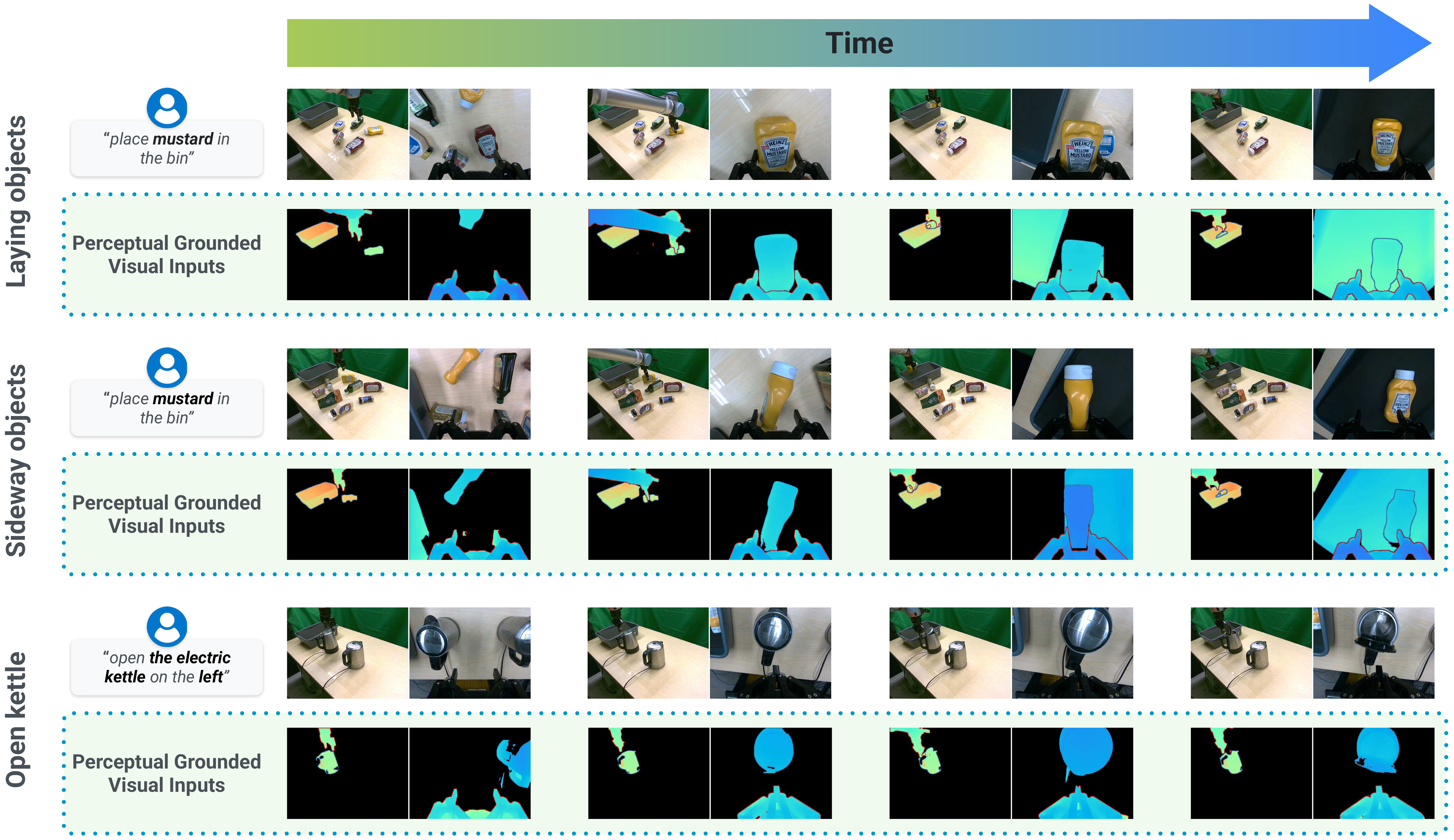}
    \caption{\edit{\textbf{Qualitative results in additional evaluations.} Top two rows: toppled-object pick-and-place, where the queried mustard bottle appears in lying-down and sideways poses. Bottom row: open/close evaluation, where the policy opens the left kettle while ignoring a visually similar kettle. For each example, we show the original observations and the corresponding perception-grounded visual inputs over time.}}
    \label{fig:qualitative_appendix}
\end{figure*}

\subsection{\edit{Additional Clutters: Toppled Objects}}
\label{app:toppled_objects}

\edit{\textbf{Additional demonstrations.} To extend our evaluation beyond upright grocery objects, we also study cluttered manipulation when objects are toppled. We collect 200 additional demonstrations per object for three seen objects (\emph{olive oil}, \emph{ketchup}, and \emph{mustard}), with 100 demonstrations for the object lying on its back and 100 demonstrations for the object lying sideways, for a total of 600 episodes. These 600 toppled-object episodes are appended to the upright-object demonstrations used in the main paper for training.}

\edit{\textbf{Training setup.} For this study, we focus on \modelPi, since its performance trend is broadly consistent with that of \modelFAST in the main experiments. We re-train the Pi-0 policy on the combined dataset above, where all observations are first grounded by our perception grounding module. Throughout this process, the grounding module remains fully frozen, and only the downstream VLA policy is updated. This isolates whether the same frozen grounding pipeline can support toppled-object manipulation once the action policy is exposed to the new object poses.}

\edit{\textbf{Evaluation on object distractors.} We evaluate \modelPi on the same distractor-object protocol used in the main paper, with 0, 1, 4, and 7 distractors. This is the most relevant evaluation for this study, since it directly probes clutter robustness under the additional challenge of toppled object poses. In this setting, all objects are randomly placed either sideways or on their back. For context, \Cref{tab:toppled_clutter} also includes the upright-object results from the main paper.}

\begin{table}[h]
  \centering
  \caption{\edit{Upright vs. toppled-object cluttered manipulation.}}
  \label{tab:toppled_clutter}
  \footnotesize
  \setlength{\tabcolsep}{8pt}
  \renewcommand{\arraystretch}{1.1}
  \begin{tabular}{ccccc}
    \toprule
    \edit{\textbf{Setting}} & \edit{\textbf{0 Distr.}} & \edit{\textbf{1 Distr.}} & \edit{\textbf{4 Distr.}} & \edit{\textbf{7 Distr.}} \\
    \midrule
    \edit{\textbf{Upright}} & \edit{\(93.8 \pm 9.3\)} & \edit{\(92.7 \pm 4.4\)} & \edit{\(85.0 \pm 6.9\)} & \edit{\(79.2 \pm 9.4\)} \\
    \edit{\textbf{Toppled}} & \edit{\(93.6 \pm 3.5\)} & \edit{\(89.3 \pm 4.1\)} & \edit{\(87.8 \pm 7.5\)} & \edit{\(73.9 \pm 8.1\)} \\
    \bottomrule
  \end{tabular}
\end{table}

\edit{Overall, \modelPi remains fairly robust even when objects are toppled. Compared with the upright setting, success is nearly unchanged under 0, 1, and 4 distractors, indicating that the frozen perception grounding module still transfers well to these new object poses. A slight drop appears in the 7-distractor setting (about 5 percentage points), where the toppled-object configuration likely increases occlusion and produces more frequent overlap among instances. Qualitative examples in the top two rows of \cref{fig:qualitative_appendix} further show that the grounded inputs preserve the queried object across lying-down and sideways poses.}

\subsection{\edit{Additional Task: Open/Close}}

\edit{\textbf{Task setting \& training demonstrations.} To further test whether the same perception-grounded framework can be applied beyond pick-and-place tasks, we extend the evaluation to the manipulation task of open/close. We collect a separate set of demonstrations using an electric kettle. In each demonstration, a single kettle is placed at either the left or right side of the workspace. The robot is instructed to perform one of two actions: opening the kettle by pressing the open toggle, or closing the kettle by pressing down on the lid. We collect 150 demonstrations for each location-action setting, yielding 600 demonstrations in total.}

\edit{\textbf{Training setup.} Following the toppled-object study in Appendix~\ref{app:toppled_objects}, we use \modelPi as the representative grounded policy. We fine-tune the pretrained Pi-0 policy only on this open/close demonstration set, keeping it separate from the pick-and-place data used in the main experiments. All observations are processed by the same perception grounding module before policy fine-tuning; the grounding module remains fully frozen, and only the downstream VLA policy is updated. This setting further examines the generalizability of the grounding module beyond grocery pick-and-place, covering a different object category and press-based open/close actions.}

\edit{\textbf{Evaluation protocol.} We evaluate \modelPi in a cluttered setting with two electric kettles. Because the kettles share similar appearances, this evaluation uses spatially relational instructions to specify the target. At test time, two kettles are placed simultaneously on the tabletop, one on the left and one on the right. Each instruction specifies both the target location and the desired action, requiring the policy to open or close only the queried kettle while ignoring the other kettle. We evaluate four scenarios--open left, close left, open right, and close right--with 20 real-robot rollouts per scenario. The results are summarized in \Cref{tab:kettle_open_close}.}

\begin{table}[h]
  \centering
  \caption{\edit{Open/close evaluation in a cluttered two-kettle setting.}}
  \label{tab:kettle_open_close}
  \footnotesize
  \setlength{\tabcolsep}{8pt}
  \renewcommand{\arraystretch}{1.1}
  \begin{tabular}{lcc}
    \toprule
    \edit{\textbf{Target Location}} & \edit{\textbf{Action}} & \edit{\textbf{Success Rate (\%)}} \\
    \midrule
    \edit{Left}  & \edit{Open}  & \edit{\textit{100}} \\
    \edit{Left}  & \edit{Close} & \edit{\textit{90}} \\
    \edit{Right} & \edit{Open}  & \edit{\textit{100}} \\
    \edit{Right} & \edit{Close} & \edit{\textit{100}} \\
    \bottomrule
  \end{tabular}
\end{table}

\edit{The results show that \modelPi reliably follows both the spatial target reference and the requested open/close action in this two-kettle setting. The policy achieves nearly perfect success overall, suggesting that the same perception grounding module can support a more diverse tabletop manipulation setting beyond pick-and-place, while enabling the downstream VLA to focus on the queried kettle despite the presence of a visually similar distractor. The bottom row of \cref{fig:qualitative_appendix} illustrates a representative rollout for opening the left kettle.}

\subsection{\edit{Failure-Mode Analysis}}

\begin{figure}[h]
    \centering
    \includegraphics[width=0.8\linewidth]{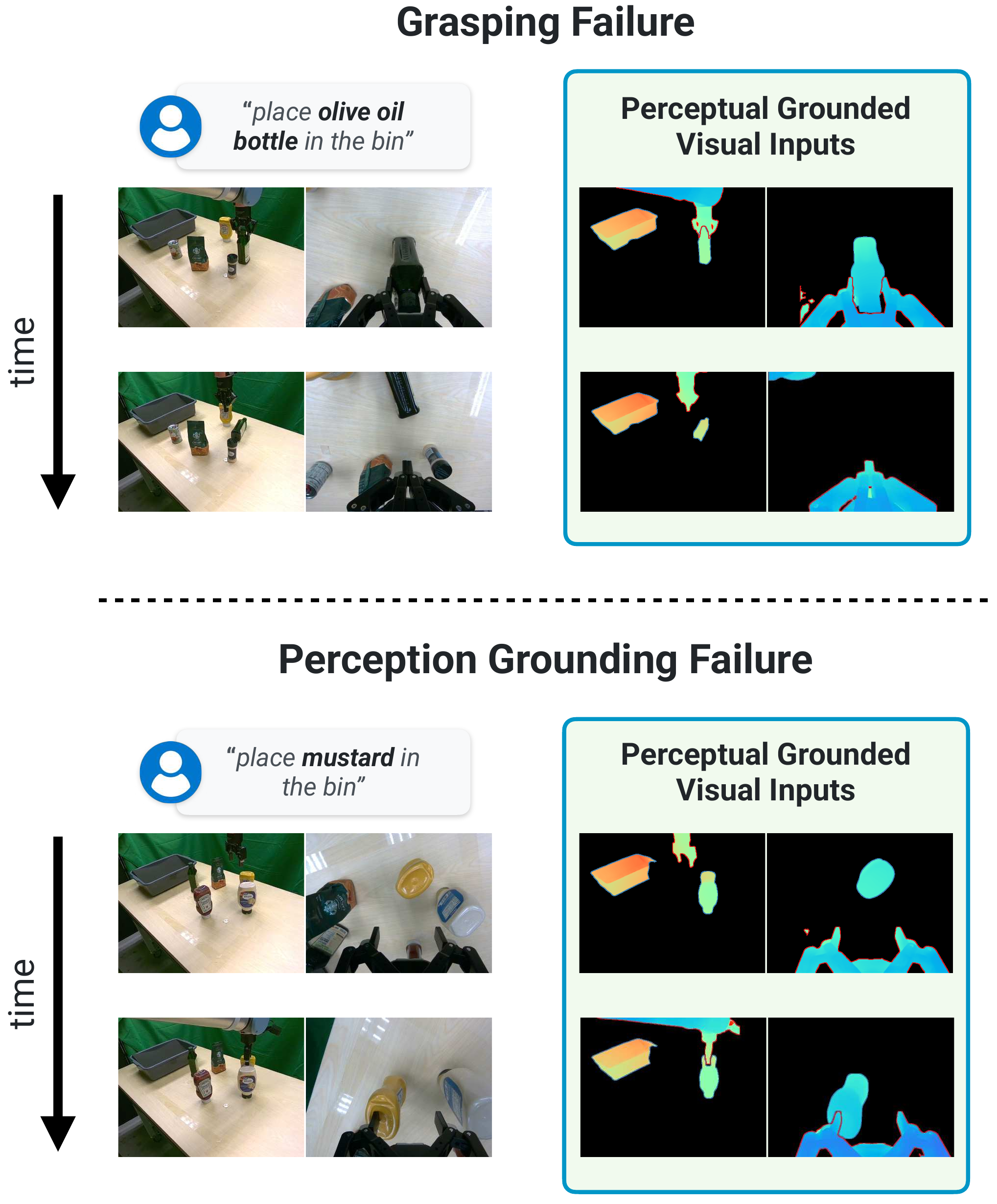}
    \caption{\edit{\textbf{Representative failure cases.} Top: correct grounding but imperfect grasp execution. Bottom: perception-grounding failure under occlusion, where the mustard bottle is merged with the occluding mayonnaise bottle in the base view.}}
    \label{fig:qualitative_failure_cases}
\end{figure}

\edit{\Cref{fig:qualitative_failure_cases} illustrates two representative failure modes observed in our real-robot rollouts. The first and more frequent case is a grasp execution failure: the perception-grounded inputs correctly isolate the queried object, but the downstream policy does not always produce a sufficiently precise grasp. In the example shown, the object slips from the gripper after an unstable grasp; in other rollouts, the gripper can also collide with the object during approach. This indicates that, even when visual grounding is correct, the final manipulation outcome can still be limited by the precision and robustness of the learned action policy during contact-rich grasping.}

\edit{The second, less frequent case is a perception-grounding failure caused by severe occlusion. In the example, a mayonnaise bottle occludes the queried mustard bottle in the base view, causing the grounded base-view input to include the occluder and corrupt the target observation. This imperfect grounding can mislead the downstream policy and produce an imprecise approach, resulting in the gripper colliding with the target object instead of forming a stable grasp. Although the wrist view still localizes the mustard correctly, the current policy has no explicit mechanism to adaptively rely on the cleaner wrist-view input as the gripper approaches the target while suppressing the defective base-view grounding.

Taken together, these examples clarify the current fault-tolerance boundary of \model. The proposed grounding substantially reduces distractor-induced target-selection errors, but severe occlusion and high-precision grasp execution remain challenging and motivate future work on policies that are more tolerant to imperfect grounded observations.}

\end{document}